\documentclass[11pt]{article}

\usepackage{acl}

\usepackage{times}
\usepackage{latexsym}
\usepackage[utf8]{inputenc}
\usepackage[T1]{fontenc}
\usepackage{booktabs}
\usepackage{amsfonts}
\usepackage{amsmath}
\usepackage{amssymb}
\usepackage{nicefrac}
\usepackage{microtype}
\usepackage{graphicx}
\usepackage{xcolor}
\usepackage{multirow}
\usepackage{subcaption}
\usepackage{CJKutf8}
\usepackage{float}
\usepackage{placeins}

\newcommand{\chg}[1]{#1}

\title{PersonaForge: Realistic Multi-Turn User Simulation for Agentic Systems}

\author{\mdseries Hanglong Lv$^{\S\P*}$, Dawei Zhu$^{\S}$, Lei Li$^{\ddagger}$, Bowen Ye$^{\S}$, Huaqiu Liu$^{\P}$, \\
Yifan Song$^{\S}$, Bofei Gao$^{\S}$, Weimin Xiong$^{\S}$, Jinhao Dong$^{\|}$, Chenhong He$^{\S}$, \\
Lingpeng Kong$^{\ddagger}$, Qi Liu$^{\ddagger}$, Tong Yang$^{\S\dagger}$, Fuli Luo$^{\P\dagger}$ \\
\small
$^{\S}$ State Key Laboratory of Multimedia Information Processing, School of Computer Science, Peking University \\
\small
$^{\P}$ LLM-Core, Xiaomi \qquad $^{\ddagger}$ The University of Hong Kong \qquad $^{\|}$ Renmin University of China \\[0.2em]
\small
\hspace*{0.5em}\texttt{lyuhanglong@stu.pku.edu.cn} \quad \texttt{yangtong@pku.edu.cn} \quad \texttt{luofuli@xiaomi.com}\hspace*{0.5em}%
}

\makeatletter
\def\@thanks{%
  \footnotetext[0]{\hspace{-1.2em}$^*$ Contribution during internship at Xiaomi LLM-Core Team.}%
  \footnotetext[0]{\hspace{-1.2em}$^\dagger$ Co-corresponding authors.}%
}
\makeatother

\begin{document}

\maketitle

\begin{abstract}
Large language models are increasingly used as agentic workflow executors, yet existing training data and benchmarks largely assume informationally complete, single-turn queries. Our analysis of 16K real-world sessions shows that 75.9\% of interactions are multi-turn, revealing a substantial gap between how users interact with agents and how such systems are trained and evaluated.
We introduce \textbf{PersonaForge}, a user simulation framework for synthesizing realistic multi-turn user--agent interactions. PersonaForge combines a four-dimensional persona space, SOUL-driven behavioral control calibrated to real-user statistics, and Reverse Deep Construction grounded in authentic seed queries. Using PersonaForge, we construct a 6.3K-record training dataset and \textbf{PersonaForge-Bench}, a manually annotated 138-task benchmark spanning over 20 professional domains with four-dimensional scoring.Experiments on Qwen3.5-27B show that PersonaForge training improves the composite score by +4.1\%, with significant gains in Task Completion (+6.0\%) and Response Quality (+6.8\%); on MiMo-V2-Flash the composite gain reaches +15.7\% with every dimension significant. Further analyses show that PersonaForge-trained agents use fewer turns and tool calls, suggesting improved interaction efficiency, while ablations confirm the contribution of SOUL components and adaptive simulation.
Together, PersonaForge and PersonaForge-Bench establish a foundation for training and evaluating agents under realistic multi-turn user interaction.
\end{abstract}
\section{Introduction}
\label{sec:intro}
\begin{figure*}[t]
    \centering
    \includegraphics[width=\textwidth, height=6cm]{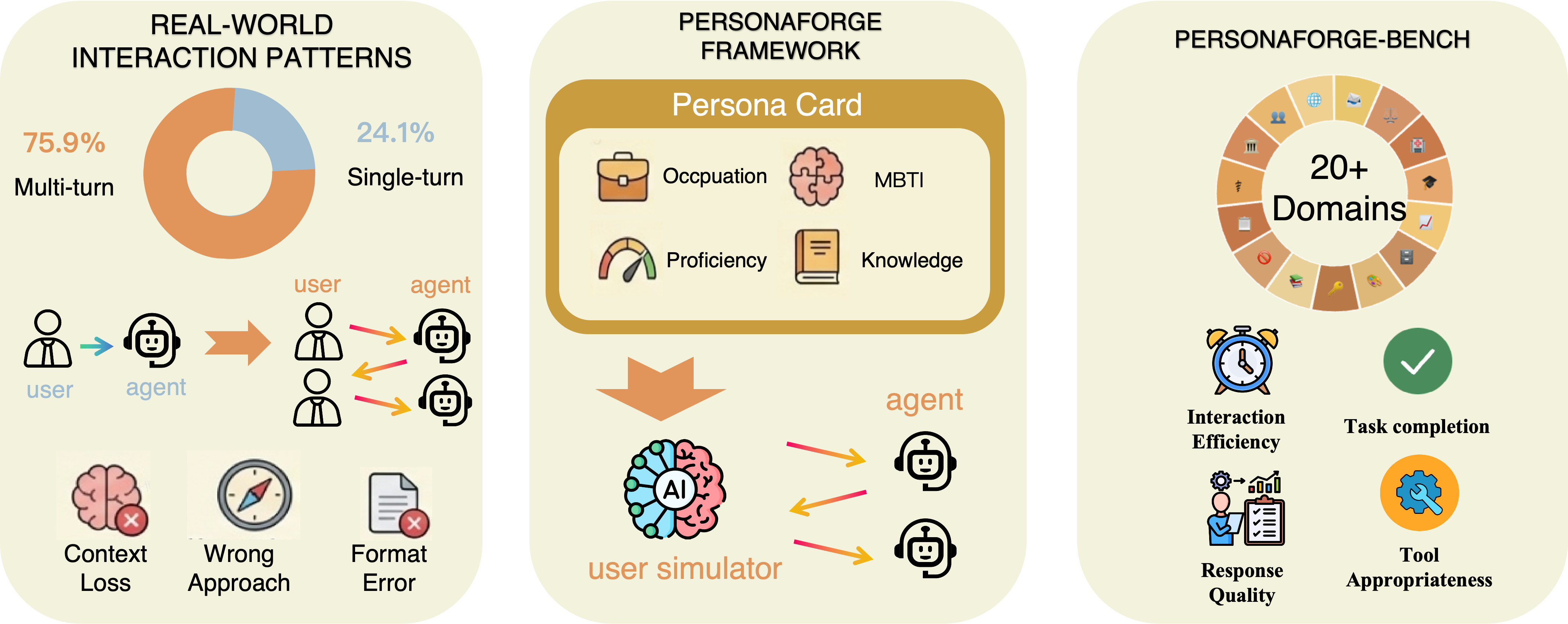}
    \caption{Overview of PersonaForge: real-session analysis, data synthesis, and benchmark evaluation.}
    \label{fig:teaser}
\end{figure*}

Large language models~\citep{ouyang2022training,openai2023gpt4} have evolved from single-turn question answerers into agentic workflow executors~\citep{yao2023react,schick2023toolformer,wu2023autogen}: systems that solve real-world tasks through tool use, multi-step reasoning, and sustained dialogue. In these settings, success depends on managing iterative, tool-intensive interactions as users steer the system over time. Yet agentic training and evaluation still often assume that the user's first message is \emph{informationally complete}. Existing tool-use data synthesis methods~\citep{liu2024toolace,liu2024apigen} typically generate offline trajectories over fixed API schemas, where a fully specified query leads to a planned tool-call chain. This misses real workflows with incomplete and evolving user intent.

Real usage looks different. Our analysis of 16K real-world sessions from public datasets (SWE-chat~\citep{baumann2026swechat} and Computer-Use~\citep{shaikh2025gum}) and internally collected sessions shows that \textbf{75.9\% of interactions are multi-turn} (Section~\ref{sec:analysis}); among multi-turn sessions, the median number of user turns is 10 and the average number of tool calls is 99. Users add requirements, change direction, inspect intermediate outputs, and correct errors; in 38.7\% of multi-turn sessions, they explicitly request corrections. This gap between complete-query assumptions and incomplete, user-steered interaction is a central bottleneck for building capable agentic systems.

Closing this gap requires a \emph{realistic user simulator} that reproduces how humans interact with agentic systems: providing incomplete requests, progressively disclosing information, and challenging incorrect outputs. Building such a simulator poses three linked challenges: (i)~\textbf{diverse yet coherent users}, whose occupations, personalities, technical proficiency, and knowledge backgrounds shape both task goals and communication styles; (ii)~\textbf{controllable turn-level behavior}, so the simulator produces terse, incremental, and sometimes corrective messages rather than verbose, overly cooperative LLM-style dialogue; and (iii)~\textbf{grounded and scalable conversations}, which originate from authentic user needs, proceed through live interaction with an agent, and avoid low-quality data.

We address these challenges with \textbf{PersonaForge}, a data synthesis framework for realistic user simulation. PersonaForge uses LLMs to act as realistic users interacting with a deployed agentic system, as illustrated in Figure~\ref{fig:teaser}. Its design maps directly to the three challenges: (1)~a \textbf{4-dimensional persona space} spanning occupation, personality, technical proficiency, and knowledge background for diverse yet coherent users; (2)~\textbf{SOUL-driven behavioral control}, which constrains messages to be terse, incremental, and corrective when needed; and (3)~\textbf{Reverse Deep Construction}, which infers persona profiles from real seed queries to ground each live interaction in authentic user needs. Applied at scale, PersonaForge produces 6.3K realistic conversations that pass quality filtering for training. To evaluate models under realistic user interaction, we further introduce \textbf{PersonaForge-Bench}, a human-verified benchmark derived from real user requests. It contains 138 tasks that reflect real user scenarios across 20+ professional domains and are designed around progressive information disclosure, embedded contradictions, and domain-specific rubrics, requiring models to elicit missing constraints, maintain state as requirements evolve, and respond appropriately to user steering. Each task requires sustained multi-turn interaction and is scored over 3 independent trials along four dimensions: interaction efficiency, tool appropriateness, task completion, and response quality, providing a controlled testbed for measuring agent behavior beyond complete-prompt execution.

\chg{Experiments on open-source models Qwen3.5-27B~\citep{qwen2025qwen3} and MiMo-V2-Flash~\citep{xiaomi2026mimov2flash} show that PersonaForge training improves the composite score across all four dimensions: +4.1\% on Qwen3.5-27B and +15.7\% on MiMo-V2-Flash, with the largest gains on task completion and response quality.} These results show training on realistic multi-turn interaction data helps models learn more effective dialogue management and tool-use strategies, strengthening their ability to interact with real users. Together, PersonaForge and PersonaForge-Bench provide a data synthesis and evaluation foundation for studying user simulation in agentic systems, moving the field beyond complete-query benchmarks toward realistic user-agent interaction.

\section{Multi-Turn Interactions in the Wild}
\label{sec:analysis}

Our analysis combines public benchmarks of agentic interaction with internal deployment logs, balancing external authority with domain coverage. The public sources are \textbf{SWE-chat} \citep{baumann2026swechat} and the \textbf{computer-use traces} of \citet{shaikh2025gum}. We complement these with conversational sessions and task-execution trajectories from an internally deployed agentic assistant, with the internal logs sampled under opt-in consent and anonymized at collection time. In aggregate, the analysis spans \textbf{16K agentic sessions}: the two public datasets each contribute 25\% and anchor reproducible baselines, while internal logs contribute 50\% and extend coverage to broader professional domains and deployed tool-mediated workflows.

\subsection{Query and Workflow Patterns}

\begin{figure}[t]
    \centering
    \includegraphics[width=\columnwidth]{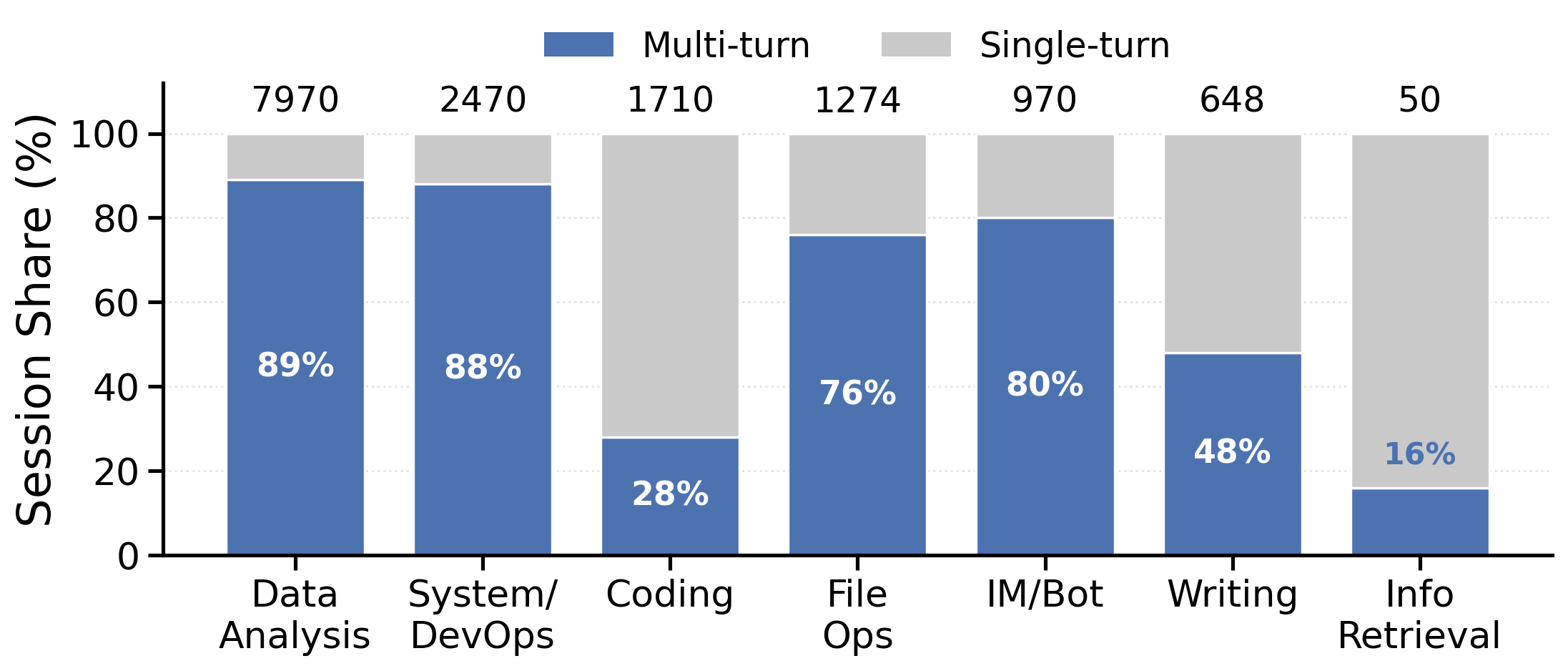}
    \caption{Task composition by category.}
    \label{fig:task_type_multiturn}
\end{figure}

\noindent\textbf{Real agentic use is heterogeneous and situated.} Figure~\ref{fig:task_type_multiturn} reports task volume and multi-turn share across seven categories, identified by keyword matching on the first user message (Appendix~\ref{sec:analysis_details}). The distribution is broad: data analysis and system/DevOps operations account for the largest volume, while coding, file operations, IM/bot interactions, writing, and information retrieval all appear as recurring use cases. These categories presuppose different user backgrounds and communication norms. A DevOps request often assumes infrastructure context and command-line fluency; a writing request may rely on style preferences and implicit audience constraints; an information-retrieval request may be issued by a user with little technical context. This heterogeneity suggests that a realistic simulator cannot sample tasks independently from users. The user profile must be coherent with the task, the domain vocabulary, and the level of expertise expressed in the conversation.

\begin{figure}[t]
    \centering
    \includegraphics[width=\columnwidth]{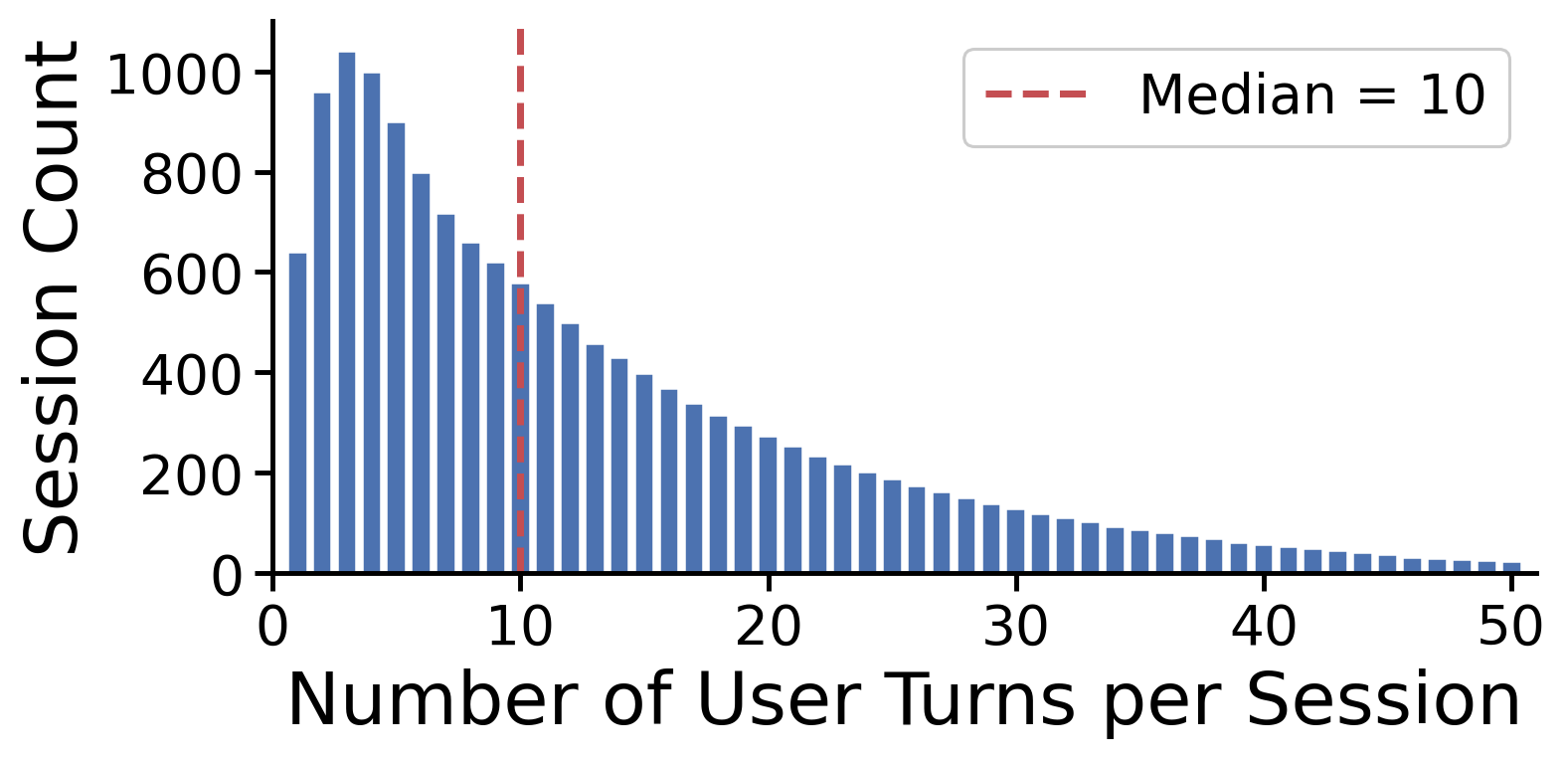}
    \caption{User turns per session.}
    \label{fig:turn_dist}
\end{figure}

\noindent\textbf{Intent is disclosed over time.} Aggregating across categories, Figure~\ref{fig:turn_dist} shows that \textbf{75.9\%} of sessions involve two or more user turns, with a median of 10 user turns. Real users therefore rarely provide a complete task specification upfront. Instead, they add constraints, inspect intermediate results, clarify ambiguous requirements, and revise the request as the assistant acts. This contrasts sharply with the training corpus underlying our deployed assistant, which is \emph{100\% single-turn}, and points to a need for simulated users that can reveal intent incrementally rather than only issue complete initial queries.

\subsection{Execution-Coupled Feedback}

\begin{table}[t]
    \centering
    \footnotesize
    \setlength{\tabcolsep}{3.5pt}
    \begin{tabular}{lccc}
        \toprule
        \textbf{Metric} & \textbf{Single} & \textbf{Multi} & \textbf{Ratio} \\
        \midrule
        Sessions & 3.9K & 12.1K & --- \\
        Avg.\ user turns & 1.0 & 25.8 & 25.8$\times$ \\
        Avg.\ tool calls & 9.6 & 99.3 & 10.3$\times$ \\
        Avg.\ length (chars) & 45K & 212K & 4.7$\times$ \\
        User correction & 2.0\% & 38.7\% & 19.4$\times$ \\
        Post-tool corr.\ share & --- & 36.0\% & --- \\
        \bottomrule
    \end{tabular}
    \caption{Single-turn vs.\ multi-turn session statistics.}
    \label{tab:single_vs_multi}
\end{table}

\begin{figure}[t]
    \centering
    \includegraphics[width=\columnwidth]{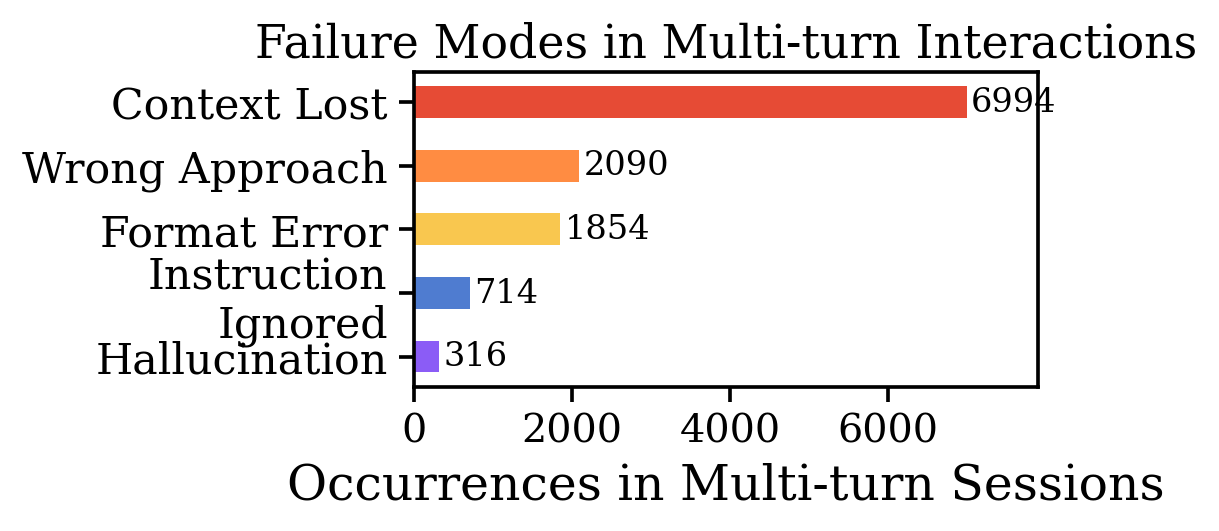}
    \caption{Failure modes in user correction messages.}
    \label{fig:failure_modes}
\end{figure}

\begin{figure*}[t]
    \centering
    \includegraphics[width=\textwidth, height=5.5cm]{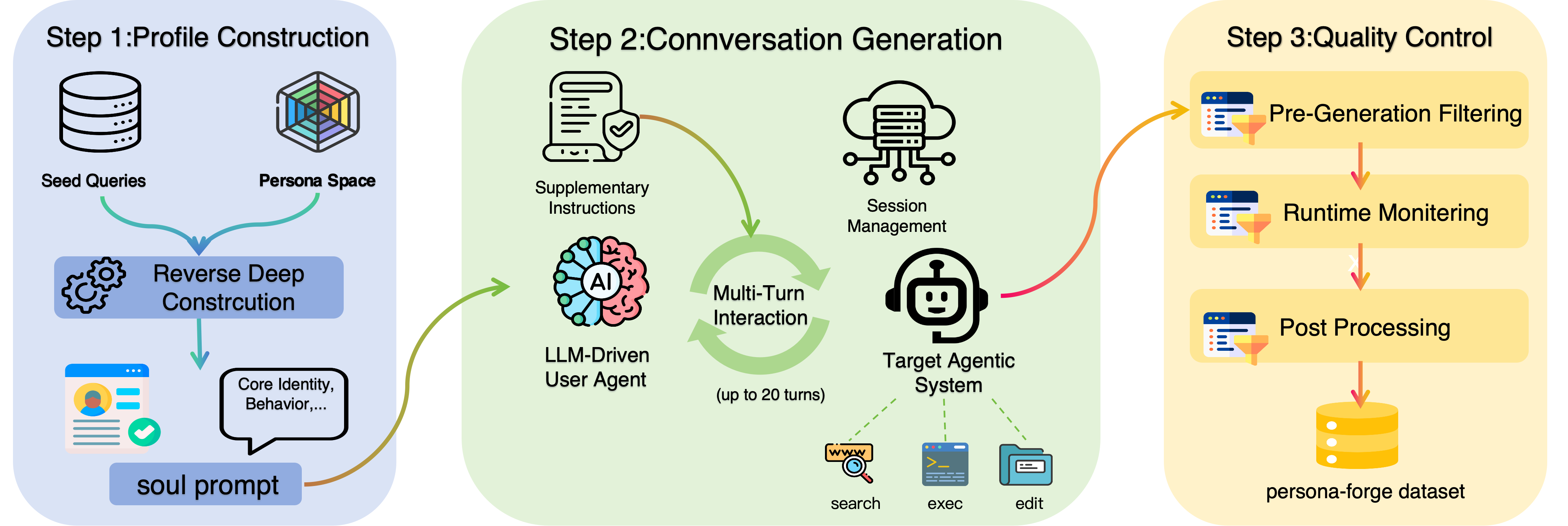}
    \caption{Overview of the PersonaForge framework.}
    \label{fig:pipeline}
\end{figure*}

\noindent\textbf{User turns are coupled to tool execution.} Table~\ref{tab:single_vs_multi} shows that multi-turn sessions contain many more tool calls (99.3 vs.\ 9.6 on average), more intermediate content (212K vs.\ 45K characters), and more user turns than single-turn sessions. We interpret these quantities not as a generic difficulty score, but as evidence that real conversations are grounded in execution. User messages respond to files, command outputs, retrieved information, partial calculations, and previous assistant decisions. This makes offline scripting insufficient: a realistic trajectory should unfold through interaction with a running agentic system, so that user follow-ups are conditioned on what the assistant actually does.

\noindent\textbf{User feedback is corrective rather than merely conversational.} Users explicitly signal errors in 38.7\% of multi-turn sessions under our correction lexicon (19.4$\times$ single-turn), and 36.0\% of corrections occur after tool execution (Appendix~\ref{sec:analysis_details}). Figure~\ref{fig:failure_modes} shows that these corrections target \emph{context loss} (7.0K), \emph{wrong approach} (2.1K), \emph{format errors} (1.9K), \emph{instruction ignored} (714), and \emph{hallucination} (316). These are not generic follow-up questions; they are closed-loop responses to the assistant's behavior. A simulator must therefore control turn-level behavior so that it can produce terse, incremental, and corrective messages when the interaction state calls for them, instead of defaulting to verbose and overly cooperative dialogue.

\subsection{Implications for User Simulation}

The analyses above portray real agentic interaction as a user-steered workflow rather than a completed query followed by execution. Three properties are especially important for realistic simulation. First, users and tasks are coupled: the same surface task can be expressed differently depending on occupation, technical proficiency, personality, and domain knowledge. Second, intent unfolds over turns: users reveal constraints, react to intermediate outputs, and correct the assistant when needed. Third, the conversation is grounded in the assistant's actual behavior, especially tool calls and partial results. These observations motivate us to design a data-generation framework that controls these axes explicitly: who the user is, how the user communicates at each turn, and what authentic task context grounds the interaction.

\section{PersonaForge}
\label{sec:framework}

PersonaForge turns a real seed query into a trainable multi-turn user--agent trajectory. A realistic simulator cannot simply sample a persona or ask a fully specified prompt: it must keep the user compatible with the task, decide what hidden intent to reveal at each turn, and ground follow-ups in the assistant's actual execution. PersonaForge therefore decomposes synthesis into three stages that correspond to these requirements. \textbf{Profile construction} (\S\ref{sec:construction}) defines who the user is and why the task is natural for them. \textbf{Conversation generation} (\S\ref{sec:soul}) controls how the user reveals intent, reacts, and corrects the assistant over turns. \textbf{Quality control} (\S\ref{sec:qc}) filters live interaction traces so that only coherent, user-like trajectories enter the retained corpus (Figure~\ref{fig:pipeline}).

\subsection{Profile Construction}
\label{sec:construction}
\label{sec:persona}

Profile construction addresses user--task coherence by defining the hidden state from which the simulated user will act. Rather than sampling tasks and users independently, PersonaForge links them through a real seed query: the query anchors an authentic task need, while the persona space determines the user's background, expertise, and communication style. We instantiate this state from a 4-dimensional persona space. \textbf{Occupation} (70 types) spans tech workers, professionals, students, entrepreneurs, and more, each annotated with tech affinity and compatible knowledge areas. \textbf{Personality type} (16 MBTI types) determines communication style; for example, INTJ users are terse and efficiency-focused, while ENFP users are curious and prone to topic shifts. \textbf{Technical proficiency} (3 levels: expert, intermediate, novice) controls whether the user speaks in jargon or plain language. \textbf{Knowledge background} (25 domains) encodes domain expertise with structured fields for expert skills, familiar topics, and blind spots. To keep these dimensions internally consistent, we enforce compatibility rules based on annotated affinities before generation; for example, a ``non-technical'' user is not paired with a high-tech-affinity occupation (see Figure~\ref{fig:coherence_example} in the Appendix).

\noindent\textbf{Reverse Deep Construction.} Given the persona space, we construct profiles via \textbf{Reverse Deep Construction}: starting from real seed queries drawn from WildChat~\citep{zhao2024wildchat} and LMSYS-Chat-1M~\citep{zheng2024lmsyschat}, an LLM works backward to infer a plausible persona profile, then generates a task scenario and connected memories naturally grounded in that persona. This design anchors each conversation in an authentic user need, preserving real-world task distributions rather than relying on synthetically generated queries. We contrast reverse-constructed profiles with forward-sampled ones on matched cases in Appendix~\ref{sec:construction_appendix} to illustrate the resulting differences in anchor specificity and connected-memory depth. 

\noindent\textbf{SOUL prompt structure.} The SOUL prompt is the executable representation of this user state. It specifies a simulated user's identity, communication style, and hidden context, adapting the structured-prompt principle used in deployed assistant systems such as OpenClaw\footnote{\url{https://openclaw.ai}} to the \emph{user} side of the interaction. The prompt has four components: (1)~a core identity paragraph grounding the persona in a specific life situation; (2)~personality--behavior mappings expressed as abstract parameters rather than utterance exemplars, which we found collapse diversity; (3)~communication style parameters (message length, formality, emoji usage); and (4)~connected memory templates that pre-establish session-specific facts, preventing the simulator from fabricating contradictory details when answering clarification questions. Connected memory is identified in our ablation as the single most impactful individual component.

\subsection{Conversation Generation}
\label{sec:soul}

Conversation generation addresses controllable turn-level behavior by converting the hidden user state into observable user messages. The key design is a deliberate \emph{information asymmetry} (see Figure~\ref{fig:interaction_structure} in the Appendix): the simulator can access the SOUL prompt, connected memories, and task scenario, while the target system sees only accumulated user messages and its own tool outputs. This asymmetry lets the simulator reveal intent gradually instead of providing a complete task specification at the first turn. Each turn, the simulator conditions on the hidden state and on what the target system has actually done, producing short, naturalistic messages that add requirements, introduce realistic inconsistencies, or correct the assistant when the interaction calls for it.

\noindent\textbf{Behavioral rules.} Beyond the persona-specific content, every SOUL prompt includes fixed behavioral rules that make the simulator behave like a user rather than another assistant: \emph{message brevity} (short follow-ups rather than full task restatements), \emph{one question per turn} to avoid unnatural multi-question patterns, \emph{active verification} of numerical results and code outputs to replicate real error correction behavior, \emph{no AI self-awareness} to maintain persona consistency, and \emph{natural termination} via a signal when the task is complete.

\subsection{Quality Control}
\label{sec:qc}

Quality control addresses scalable data construction by preventing common simulator failures from entering the training corpus. Because PersonaForge generates data through LLM-driven live interaction, a small number of role or format failures can otherwise contaminate many long trajectories. The filters therefore operate at three stages.
\emph{Pre-generation}: persona combinations are checked for coherence and seed queries are filtered for minimum complexity.
\emph{Runtime}: we detect \emph{role reversal}, where the simulator shifts from issuing directives to completing the task itself, using a character-length heuristic that flags user messages exceeding 1{,}500 characters (calibrated on real-user data where 99.2\% of follow-ups fall below this cutoff; flagged messages are additionally checked for code-paste patterns to avoid false positives).
\emph{Post-processing}: conversations undergo role-reversal truncation, degenerate pattern filtering, format normalization, and tool-configuration randomization.

\label{sec:dataset_stats}

\noindent\textbf{Retained corpus.} Table~\ref{tab:dataset_stats} summarizes the trajectories retained after filtering and formatting for training. The final data contains roughly 6.3K session-level records and 430K messages, with 96\% being multi-turn (mean 9.3 user turns). User messages remain short, with a median length of about 100 characters. Detailed persona dimension distributions and quality control pipeline statistics are provided in Appendix~\ref{sec:persona_detail}.

\begin{table*}[t]
  \centering
  \small
  \begin{tabular}{lr|lr}
    \toprule
    \textbf{Statistic} & \textbf{Value} & \textbf{Persona Dimension} & \textbf{Unique / Norm.\ H} \\
    \midrule
    Sessions  & 6.3K & Occupation & 70 / 0.79 \\
    Messages (user / asst / tool) & 59K / 103K / 268K & MBTI & 16 / 0.77 \\
    Mean user turns & 9.3 & Tech Level & 3 / 0.93 \\
    User msg length (median) & 100 chars & Knowledge BG & 25 / 0.91 \\
    \bottomrule
  \end{tabular}
  \caption{Retained training corpus statistics. \emph{Norm.\ H} is normalized Shannon entropy.}
  \label{tab:dataset_stats}
\end{table*}

\section{PersonaForge-Bench}
\label{sec:benchmark}

\noindent\textbf{PersonaForge-Bench targets realistic user-agent interaction.} Section~\ref{sec:intro} identifies three requirements for realistic user simulation: diverse yet coherent users, controllable turn-level behavior, and conversations grounded in authentic needs and live agent behavior. Section~\ref{sec:analysis} shows why these requirements matter in practice: real agentic use is heterogeneous across domains, user intent is disclosed over time, and feedback is often corrective and coupled to tool execution. A benchmark built from complete single-turn prompts would hide precisely these behaviors. We therefore construct 138 tasks spanning 20+ professional domains, with scenarios in both Chinese and English, on top of \textsc{Claw-Eval}~\citep{ye2026claweval}. Each task carries a fine-grained domain tag, and these tags are aggregated into 8 reporting buckets; we release the keyword priority lexicon and the full task$\to$tag$\to$bucket mapping. \chg{Domain composition, bucket assignments, and the mapping mechanism are detailed in Appendix~\ref{sec:rubric_examples} and~\ref{sec:per_domain_detail}.} Each task is derived from a realistic user scenario and requires sustained multi-turn interaction rather than one-shot task execution.

\noindent\textbf{The benchmark faithfully evaluates user--LLM interaction trajectories.} Its tasks are built around three principles that correspond to the observations above: \emph{progressive information disclosure} tests whether the assistant can ask for and integrate missing constraints; \emph{embedded contradictions} test whether it can maintain state and detect conflicts as new information arrives; and \emph{domain-specific rubrics} ensure that success is judged against the concrete requirements of the professional scenario rather than generic helpfulness. We score each task along four equally weighted dimensions: Interaction Efficiency, Tool Appropriateness, Task Completion, and Response Quality (detailed definitions in Appendix~\ref{sec:rubric_examples}). These dimensions match the failure surfaces observed in Section~\ref{sec:analysis}: efficiency and tool appropriateness measure management of tool-intensive workflows, task completion measures recovery of hidden requirements and contradictions, and response quality measures whether the final answer is accurate and structured. Scoring is \emph{task-specific}: each task includes a rubric with required clarification items, reference calculations with tolerance windows, and evaluation chains, ensuring the LLM judge (DeepSeek-V4-Pro~\citep{deepseekai2026deepseekv4pro}) evaluates against domain-appropriate criteria. The composite score averages all four dimensions across 3 independent trials per task, and human validation ($\kappa_{jh} = 0.79$, $r = 0.84$; Appendix~\ref{sec:judge_robustness}) confirms near-human reliability.

\section{Experiments}
\label{sec:experiments}

This section evaluates whether PersonaForge data improves agentic interaction behavior and whether the gains hold beyond the PersonaForge-Bench setting. We first compare PersonaForge-trained models with their base counterparts and strong external systems, then analyze robustness, data-source effects, SOUL ablations, and transfer to an independent agent benchmark.

\begin{table*}[t]
    \centering
    \footnotesize
    \begin{tabular}{lccccc}
        \toprule
        \textbf{Model} & \textbf{Eff.} & \textbf{Tool} & \textbf{Compl.} & \textbf{Resp.} & \textbf{Overall} \\
        \midrule
        Claude Opus 4.6   & 83.0\% & 95.3\% & 86.5\% & 87.8\% & 88.2\% \\
        Gemini 3.1 Pro    & 83.8\% & 88.0\% & 85.0\% & 84.5\% & 85.9\% \\
        GPT-5.5           & 81.0\% & 93.5\% & 86.2\% & 87.0\% & 86.8\% \\
        \midrule
        MiMo-V2-Pro       & 81.7\% & 91.7\% & 84.6\% & 84.7\% & 85.7\% \\
        Kimi-K2.5         & 80.0\% & 87.8\% & 81.4\% & 85.0\% & 83.6\% \\
        MiniMax-M2.7      & 76.4\% & 88.3\% & 79.7\% & 78.0\% & 80.6\% \\
        GLM-5.1           & 73.6\% & 85.5\% & 70.3\% & 72.5\% & 75.5\% \\
        \midrule
        Qwen3.5-27B (Base)       & 74.0\% & 88.1\% & 70.6\% & 72.1\% & 76.2\% \\
        Qwen3.5-27B + PersonaForge   & 76.5\% & 89.1\% & 76.5\% & 78.9\% & 80.3\% \\
        \midrule
        MiMo-V2-Flash (Base)     & 53.2\% & 73.5\% & 53.5\% & 61.3\% & 60.4\% \\
        MiMo-V2-Flash + PersonaForge & 69.8\% & 83.7\% & 75.5\% & 75.4\% & 76.1\% \\
        \bottomrule
    \end{tabular}
    \caption{\chg{Overall PersonaForge-Bench performance. Paired bootstrap 95\% CIs and significance markers for the PersonaForge deltas are reported in Table~\ref{tab:ablation_summary} (Qwen3.5-27B) and Table~\ref{tab:flash_ci} (MiMo-V2-Flash, Appendix~\ref{sec:significance}).}}
    \label{tab:leaderboard}
\end{table*}

\subsection{Setup}
\label{sec:setup}

\noindent\textbf{Model and training.} We fine-tune \textbf{Qwen3.5-27B}~\citep{qwen2025qwen3} and \textbf{MiMo-V2-Flash}~\citep{xiaomi2026mimov2flash}, covering both a 27B-scale model and a larger-scale MoE model, with full-parameter supervised fine-tuning (SFT) on the 6.3K conversations retained from the PersonaForge pipeline. Both models are trained on NVIDIA A100 GPUs for 5 epochs with a learning rate of $5\times10^{-5}$ and a 128K maximum context length; training takes 2{,}550 GPU-hours total.

\noindent\textbf{Evaluation protocol.} We evaluate on the full 138 tasks of PersonaForge-Bench. For each task, we run 3 independent trials; all trials are scored and the final score is the average across trials and tasks. The user simulator is Claude Opus 4.6~\citep{anthropic2026opus46}; the LLM judge is DeepSeek-V4-Pro~\citep{deepseekai2026deepseekv4pro} with the task-specific scoring protocol described in Section~\ref{sec:benchmark}. We additionally evaluate transfer on held-out tasks from \textsc{Claw-Eval}~\citep{ye2026claweval}. We report Overall scores and per-dimension breakdowns. A full per-domain score table is provided in Appendix~\ref{sec:per_domain_detail}. We verify strict separation between training and evaluation data in Appendix~\ref{sec:data_separation}.

\noindent\textbf{Baselines.} We compare PersonaForge-trained models against their zero-shot base counterparts: Qwen3.5-27B~\citep{qwen2025qwen3} and MiMo-V2-Flash~\citep{xiaomi2026mimov2flash}. We additionally include Qwen3.5-27B trained on real-user replay data produced from our internally collected user-agent interaction logs, using the same data scale as the PersonaForge SFT corpus, isolating the effect of synthetic versus authentic interaction trajectories. As external references, we include leading closed-source models (Claude Opus 4.6~\citep{anthropic2026opus46}, Gemini 3.1 Pro~\citep{google2026gemini31pro}, GPT-5.5~\citep{openai2026gpt55}) and competitive proprietary models (MiMo-V2-Pro~\citep{xiaomi2026mimov2pro}, Kimi-K2.5~\citep{moonshot2026kimi25}, MiniMax-M2.7~\citep{minimax2026m27}, GLM-5.1~\citep{glmteam2026glm5}); all evaluated under an identical, rigorously controlled evaluation protocol.

\subsection{Main Results}
\label{sec:main_results}

\chg{Table~\ref{tab:leaderboard} reports the overall performance. On Qwen3.5-27B, the 27B-scale base model reaches a composite score of 76.2\% with strong Tool Appropriateness (88.1\%). PersonaForge yields a $+4.1\%$ composite gain, concentrated in Task Completion ($+6.0\%$) and Response Quality ($+6.8\%$), with all four dimensions improving. This shows that PersonaForge provides a consistent interaction-training signal even when the base model already has capable tool-use behavior.}

\chg{The gain is substantially larger on MiMo-V2-Flash, the larger-scale MoE model. PersonaForge improves the composite score from 60.4\% to 76.1\%, a $+15.7\%$ absolute gain (26.0\% relative), with the largest improvement on Task Completion ($+22.0\%$, +41.1\% relative). Tool Appropriateness also rises by $+10.1\%$, indicating that the additional capacity is not only reflected in final-answer quality but also in more effective use of the interaction and tool budget. Together, these results suggest a clear scaling pattern: PersonaForge improves both evaluated models, but the effect is more pronounced in the larger-parameter, large-scale MoE setting.}

\subsection{Ablation Studies}
\label{sec:ablations}

\begin{table*}[t]
    \centering
    \footnotesize
    \begin{tabular}{llccccc}
        \toprule
        \textbf{Ablation} & \textbf{Condition} & \textbf{Eff.} & \textbf{Tool} & \textbf{Compl.} & \textbf{Resp.} & \textbf{Overall} \\
        \midrule
        \multicolumn{7}{l}{\textit{Training-data effect on Qwen3.5-27B}} \\
        Qwen3.5-27B & Base & 74.0\% & 88.1\% & 70.6\% & 72.1\% & 76.2\% \\
        Qwen3.5-27B & + PersonaForge & 76.5\% & 89.1\% & 76.5\% & 78.9\% & 80.3\% \\
        Qwen3.5-27B & $\Delta$ & $+2.6\%$ & $+1.0\%$ & $+6.0\%^{\chg{*}}$ & $+6.8\%^{\chg{*}}$ & $+4.1\%^{\chg{*}}$ \\
        \midrule
        \multicolumn{7}{l}{\textit{SOUL component ablation on Qwen3.5-27B}} \\
        Qwen3.5-27B & Full PersonaForge & 76.5\% & 89.1\% & 76.5\% & 78.9\% & 80.3\% \\
        Qwen3.5-27B & $-$Connected Memory & 75.5\% & 89.5\% & 73.0\% & 77.5\% & 78.9\% \\
        Qwen3.5-27B & $-$Flaws \& Traits & 74.8\% & 88.0\% & 70.5\% & 79.5\% & 78.2\% \\
        \midrule
        \multicolumn{7}{l}{\textit{User-simulation ablation on Qwen3.5-27B}} \\
        Qwen3.5-27B & Real-User Replay & 72.5\% & 88.5\% & 68.0\% & 73.0\% & 75.6\% \\
        Qwen3.5-27B & PersonaForge & 76.5\% & 89.1\% & 76.5\% & 78.9\% & 80.3\% \\
        Qwen3.5-27B & $\Delta$ & $+4.0\%$ & $+0.6\%$ & $+8.5\%$ & $+5.9\%$ & $+4.7\%$ \\
        \bottomrule
    \end{tabular}
    \caption{PersonaForge-Bench ablations on Qwen3.5-27B. \chg{$^{*}$ marks a significant delta (paired bootstrap 95\% CI excludes zero, 10,000 resamples over 138 tasks; Wilcoxon $p<.05$). For the training-data $\Delta$ row: composite 95\% CI is $[+1.9,+6.4]$ ($p=.037$), with Completion $p=.012$ and Response $p=.006$; Tool and Efficiency remain within noise and are unmarked. One cell inverts non-monotonically: $-$Connected Memory scores 89.5\% on Tool versus the full 89.1\%; with a Tool CI half-width of roughly $\pm2.5$, this 0.4-point inversion is noise and not an interpretable gain.} MiMo-V2-Flash is reported in Appendix~\ref{tab:mimo_partial_ablation}.}
    \label{tab:ablation_summary}
\end{table*}

\noindent\textbf{SOUL component ablation.} Table~\ref{tab:ablation_summary} shows that PersonaForge's behavioral controls are important for downstream performance. On Qwen3.5-27B, removing connected memory reduces the composite score from 80.3\% to 78.9\%, with the largest drop on Task Completion ($-3.5\%$). Removing flaws and traits further reduces performance to 78.2\%, indicating that both latent context and persona-specific behavioral variation contribute to the training signal. \chg{The single non-monotonic cell in the table, Tool Appropriateness rising from 89.1\% to 89.5\% when connected memory is removed, sits within the Tool noise band (CI half-width $\approx \pm 2.5$) so the ablation conclusions are not materially affected.}

\noindent\textbf{User-simulation ablation.} We also compare PersonaForge against a real-user replay baseline built from our internally collected user-agent interaction logs. Replay preserves authentic human utterances but fixes the user script, so it cannot adapt to the assistant's actual responses. PersonaForge outperforms replay by $+4.7\%$ composite, with the largest gain on Task Completion ($+8.5\%$), showing that adaptive simulation provides value beyond exposing the model to human-written multi-turn transcripts.

We further verify that these gains are robust to scoring protocol and judge choice, and are not artifacts of response length or formatting bias; full analyses are provided in Appendix~\ref{sec:judge_robustness}.

\subsection{Analysis}
\label{sec:analysis_experiments}

\noindent\chg{\textbf{Training method does not confound the gains.} LoRA trained on our corpus under the identical data scale, learning rate, and epochs reproduces the full-SFT profile, and the corpus transfers to a third-party independent benchmark. Detailed training controls and transfer results are provided in Appendix~\ref{sec:data_recipe_transfer}.}

\noindent\textbf{SOUL components contribute synergistically.} Behavioral ablations across 40 diverse personas ($N\!=\!240$ conversations) show that SOUL components jointly shape realistic user behavior (full results in Appendix~\ref{sec:soul_ablation_tables}). Connected memory anchors progressive disclosure by giving the simulator latent context to reveal over time, while flaws and traits constrain how that disclosure unfolds. This explains why removing either component weakens downstream Task Completion: the model receives less signal about how to recover hidden intent through interaction.

\noindent\textbf{Scale effects and interaction efficiency.} Gains are positive across all eight evaluated domains (see Appendix~\ref{sec:per_domain_detail}, which also describes the keyword-matching rule used for domain assignment). For MiMo-V2-Flash, the largest domain gains are Policy \& Engineering (+29.6\%) and Education \& Healthcare (+23.7\%); Qwen3.5-27B shows smaller but still broad gains, led by Policy \& Engineering (+6.0\%) and Technology \& Data (+5.3\%). Figure~\ref{fig:efficiency} quantifies the interaction-efficiency shift for the larger-scale MoE model: MiMo-V2-Flash after PersonaForge SFT uses 20.7\% fewer turns, 9.2\% fewer tool calls, and \textbf{54.2\% fewer web\_fetch calls}. Tasks reaching the 25-turn limit drop from 73 to 30 ($-58.9\%$), while Task Completion increases by $+22.0\%$. Qwen3.5-27B follows the same direction: turns drop from 17.8 to 14.3 ($-19.7\%$), tool calls from 18.5 to 13.8 ($-25.5\%$), and 25-turn-limit cases from 133 to 53 ($-60.2\%$), while Task Completion increases by $+6.0\%$. \chg{This shows that PersonaForge supervision improves both completion quality and interaction selectivity, with the effect most pronounced on the model whose base starts lower and therefore has more headroom.}

\noindent\chg{\textbf{Gains generalize to out-of-distribution tasks.} On held-out tasks from \textsc{Claw-Eval} (no overlap with PersonaForge-Bench), MiMo-V2-Flash improves from 59.0\% to 69.1\% ($+10.1\%$). Two threshold-based metrics move in the same direction: Pass@3, the fraction of tasks where at least one of the 3 trials scores $\ge 75$, rises from 52.6\% to 76.3\%; Pass$^3$, the stricter fraction where all 3 trials score $\ge 75$, rises from 10.5\% to 23.7\%. Qwen3.5-27B shows more modest gains (68.9\%$\to$70.3\%, $+1.4\%$), mirroring its smaller PersonaForge-Bench improvement. These out-of-distribution results confirm that PersonaForge instills general interaction competencies rather than overfitting to benchmark rubrics.}

\section{Related Work}
\label{sec:related}

\noindent\textbf{Agentic LLMs.} Recent LLM systems increasingly act as agents that reason, call tools, and operate in external environments. ReAct~\citep{yao2023react} couples reasoning traces with actions, Toolformer~\citep{schick2023toolformer} studies self-supervised tool use, and AutoGen~\citep{wu2023autogen} builds multi-agent workflows through conversation. Evaluation has likewise shifted from static QA to executable agent tasks: AgentBench~\citep{liu2023agentbench} evaluates agents across interactive environments; SWE-bench~\citep{jimenez2024swebench} tests real GitHub issue resolution; WebArena~\citep{zhou2023webarena} and OSWorld~\citep{xie2024osworld} evaluate web and desktop operation; BrowseComp~\citep{wei2025browsecomp} focuses on persistent web browsing; and $\tau$-bench~\citep{yao2024taubench} studies tool-agent-user interaction in constrained service domains. These benchmarks establish the importance of long-horizon tool use, but most evaluate agents against fixed tasks rather than training and evaluating them under realistic, progressively steering users.

\noindent\textbf{Persona modeling and user simulation.} Persona-based generation has become a common mechanism for diversifying synthetic behavior. PersonaHub~\citep{chan2024personahub} curates large-scale persona inventories, while \citet{li2025llmpersona} show that LLM-generated personas can introduce systematic biases. Generative Agents~\citep{park2023generative} and CAMEL~\citep{li2023camel} combine personas, memory, and role play for sandbox simulation, and domain-specific simulators model recommendation feedback~\citep{cai2025agentic} or personality-driven dynamics~\citep{kroczek2025influence,depaula2025effects}. PersonaForge differs by using persona modeling not only to vary user identity, but also to control communication style, hidden context, progressive disclosure, and corrective behavior during live interaction with a deployed agentic system.

\section{Conclusion}
\label{sec:conclusion}

We presented PersonaForge, an effective framework for synthesizing realistic multi-turn user interactions, and PersonaForge-Bench, a benchmark for evaluating agentic systems under progressive user steering. Training on PersonaForge data consistently improves model performance across dense and MoE architectures and across model scales, and the improvements transfer to held-out agentic tasks beyond PersonaForge-Bench. These experiments suggest that realistic user simulation is a practical path for studying and improving agentic interaction, offering valuable insight into how models learn to manage incomplete intent, correction, and tool-intensive workflows.

\section*{Limitations}

\noindent\textbf{Persona design choices.} We adopt MBTI types as the personality dimension not for psychometric validity but for practical controllability: the 16 discrete types provide an interpretable, well-documented space for steering communication style (e.g., INTJ$\to$terse, ENFP$\to$exploratory), and LLMs respond reliably to MBTI-based prompting. We do not claim that MBTI captures real personality structure; it serves as a behavioral diversity mechanism. Alternative trait models (e.g., Big Five) could substitute with similar diversity effects.

\noindent\textbf{Ethical considerations.} All interaction logs analyzed in Section~\ref{sec:analysis} were collected with explicit opt-in consent from participants, who were informed that anonymized usage patterns would be used for research. No individual sessions are reproduced in the paper. For medical and legal benchmark tasks, all scenarios are fictional composites that do not correspond to real patients or cases, and model outputs are not intended as professional advice.

\section*{Acknowledgments}

\chg{This work was supported by the National Key Research and Development Program of China under Grant No. 2024YFB2906603, and in part by the National Natural Science Foundation of China (NSFC) (No. 62372009).}

\bibliography{references}

\begin{thebibliography}{43}
\providecommand{\natexlab}[1]{#1}

\bibitem[{{Anthropic}(2026)}]{anthropic2026opus46}
{Anthropic}. 2026.
\newblock \href
  {https://www-cdn.anthropic.com/14e4fb01875d2a69f646fa5e574dea2b1c0ff7b5.pdf}
  {Claude opus 4.6 system card}.

\bibitem[{Baumann et~al.(2026)Baumann, Padmakumar, Li, Yang, Yang, and
  Koyejo}]{baumann2026swechat}
Joachim Baumann, Vishakh Padmakumar, Xiang Li, John Yang, Diyi Yang, and Sanmi
  Koyejo. 2026.
\newblock \href {https://arxiv.org/abs/2604.20779} {Swe-chat: Coding agent
  interactions from real users in the wild}.
\newblock \emph{arXiv preprint arXiv:2604.20779}.

\bibitem[{Cai et~al.(2025)Cai, Zhang, Bao, Gao, Wang, Feng, and
  He}]{cai2025agentic}
Shihao Cai, Jizhi Zhang, Keqin Bao, Chongming Gao, Qifan Wang, Fuli Feng, and
  Xiangnan He. 2025.
\newblock \href {https://arxiv.org/abs/2410.20027} {Agentic feedback loop
  modeling improves recommendation and user simulation}.
\newblock In \emph{Proceedings of the 48th International ACM SIGIR Conference}.

\bibitem[{Chan et~al.(2024)Chan, Wang, Yu, Mi, and Yu}]{chan2024personahub}
Xin Chan, Xiaoyang Wang, Dian Yu, Haitao Mi, and Dong Yu. 2024.
\newblock \href {https://arxiv.org/abs/2406.20094} {Scaling synthetic data
  creation with 1,000,000,000 personas}.
\newblock \emph{arXiv preprint arXiv:2406.20094}.

\bibitem[{de~Paula et~al.(2025)de~Paula, Culpepper, Moffat, Cherumanal,
  Scholer, and Trippas}]{depaula2025effects}
Angel Felipe~Magnoss{\~a}o de~Paula, J.~Shane Culpepper, Alistair Moffat,
  Sachin~Pathiyan Cherumanal, Falk Scholer, and Johanne Trippas. 2025.
\newblock \href {https://arxiv.org/abs/2505.11795} {The effects of demographic
  instructions on {LLM} personas}.
\newblock In \emph{Proceedings of the 48th International ACM SIGIR Conference}.

\bibitem[{{DeepSeek-AI}(2026)}]{deepseekai2026deepseekv4pro}
{DeepSeek-AI}. 2026.
\newblock \href {https://huggingface.co/deepseek-ai/DeepSeek-V4-Pro}
  {{DeepSeek-V4-Pro}: Hugging face model repository}.

\bibitem[{{GLM Team}(2026)}]{glmteam2026glm5}
{GLM Team}. 2026.
\newblock \href {https://arxiv.org/abs/2602.15763} {Glm-5: From vibe coding to
  agentic engineering}.
\newblock \emph{arXiv preprint arXiv:2602.15763}.

\bibitem[{{Google DeepMind}(2026)}]{google2026gemini31pro}
{Google DeepMind}. 2026.
\newblock \href {https://deepmind.google/models/model-cards/gemini-3-1-pro}
  {Gemini 3.1 pro model card}.

\bibitem[{Hao et~al.(2025)Hao, Zeng, Yu, Li, and Wang}]{hao2025acebench}
Xinlong Hao, Xingshan Zeng, Shuai Yu, Dexun Li, and Xinzhi Wang. 2025.
\newblock Acebench: Who wins the match point in tool usage?
\newblock In \emph{Findings of the Association for Computational Linguistics:
  EMNLP 2025}, pages 12970--12998.

\bibitem[{Hu et~al.(2022)Hu, Shen, Wallis, Allen-Zhu, Li, Wang, Wang, and
  Chen}]{hu2022lora}
Edward~J. Hu, Yelong Shen, Phillip Wallis, Zeyuan Allen-Zhu, Yuanzhi Li, Shean
  Wang, Lu~Wang, and Weizhu Chen. 2022.
\newblock Lora: Low-rank adaptation of large language models.
\newblock In \emph{International Conference on Learning Representations}.

\bibitem[{Hu and Collier(2024)}]{hu2024quantifying}
Tiancheng Hu and Nigel Collier. 2024.
\newblock Quantifying the persona effect in llm simulations.
\newblock In \emph{Proceedings of the 62nd Annual Meeting of the Association
  for Computational Linguistics}, pages 10289--10307.

\bibitem[{Jimenez et~al.(2024)Jimenez, Yang, Wettig, Yao, Pei, Press, and
  Narasimhan}]{jimenez2024swebench}
Carlos~E. Jimenez, John Yang, Alexander Wettig, Shunyu Yao, Kexin Pei, Ofir
  Press, and Karthik Narasimhan. 2024.
\newblock \href {https://arxiv.org/abs/2310.06770} {{SWE-bench}: Can language
  models resolve real-world {GitHub} issues?}
\newblock In \emph{International Conference on Learning Representations}.

\bibitem[{{Kimi Team}(2026)}]{moonshot2026kimi25}
{Kimi Team}. 2026.
\newblock \href {https://arxiv.org/abs/2602.02276} {Kimi k2.5: Visual agentic
  intelligence}.
\newblock \emph{arXiv preprint arXiv:2602.02276}.

\bibitem[{Kroczek et~al.(2025)Kroczek, May, Hettenkofer, Ruider, Ludwig, and
  M{\"u}hlberger}]{kroczek2025influence}
Leon~O.H. Kroczek, Alexander May, Selina Hettenkofer, Andreas Ruider, Bernd
  Ludwig, and Andreas M{\"u}hlberger. 2025.
\newblock \href {https://arxiv.org/abs/2411.05653} {The influence of persona
  and conversational task on social interactions with a {LLM}-controlled
  embodied conversational agent}.
\newblock \emph{Computers in Human Behavior}, 172:108759.

\bibitem[{Kyung et~al.(2025)Kyung, Chung, Bae, Kim, Sohn, Kim, Kim, and
  Choi}]{kyung2025patientsim}
Daeun Kyung, Hyunseung Chung, Seongsu Bae, Jiho Kim, Jae~Ho Sohn, Taerim Kim,
  Soo~Kyung Kim, and Edward Choi. 2025.
\newblock Patientsim: A persona-driven simulator for realistic doctor-patient
  communication.
\newblock In \emph{Advances in Neural Information Processing Systems}.

\bibitem[{Li et~al.(2025)Li, Chen, Namkoong, and Peng}]{li2025llmpersona}
Ang Li, Haozhe Chen, Hongseok Namkoong, and Tianyi Peng. 2025.
\newblock \href {https://arxiv.org/abs/2503.16527} {{LLM} generated persona is
  a promise with a catch}.
\newblock \emph{arXiv preprint arXiv:2503.16527}.

\bibitem[{Li et~al.(2023)Li, Hammoud, Itani, Khizbullin, and
  Ghanem}]{li2023camel}
Guohao Li, Hasan Abed Al~Kader Hammoud, Hani Itani, Dmitrii Khizbullin, and
  Bernard Ghanem. 2023.
\newblock \href {https://arxiv.org/abs/2303.17760} {Camel: Communicative agents
  for ``mind'' exploration of large scale language model society}.
\newblock \emph{Advances in Neural Information Processing Systems}, 36.

\bibitem[{Liu et~al.(2024{\natexlab{a}})Liu, Zeng, Cheng, Wang, Li, Liang,
  Zhao, Zhang, and Li}]{liu2024toolace}
Weiwen Liu, Xu~Zeng, Xinlong Cheng, Shuai Wang, Zishan Li, Yong Liang, Peng
  Zhao, Bo~Zhang, and Hai-Tao Li. 2024{\natexlab{a}}.
\newblock \href {https://arxiv.org/abs/2409.00920} {Toolace: Winning the points
  of llm function calling}.
\newblock \emph{arXiv preprint arXiv:2409.00920}.

\bibitem[{Liu et~al.(2023)Liu, Yu, Zhang, Xu, Lei, Lai, Gu, Ding, Men, Yang
  et~al.}]{liu2023agentbench}
Xiao Liu, Hao Yu, Hanchen Zhang, Yifan Xu, Xuanyu Lei, Hanyu Lai, Yu~Gu,
  Hangliang Ding, Kaiwen Men, Kejuan Yang, and 1 others. 2023.
\newblock \href {https://arxiv.org/abs/2308.03688} {{AgentBench}: Evaluating
  {LLMs} as agents}.
\newblock \emph{arXiv preprint arXiv:2308.03688}.

\bibitem[{Liu et~al.(2024{\natexlab{b}})Liu, Hoang, Zhang, Zhu, Lan, Kokane,
  Tan, Yao, Liu, Feng et~al.}]{liu2024apigen}
Zuxin Liu, Thai Hoang, Jianguo Zhang, Ming Zhu, Tian Lan, Shirley Kokane,
  Juntao Tan, Weiran Yao, Zhiwei Liu, Yihao Feng, and 1 others.
  2024{\natexlab{b}}.
\newblock \href {https://arxiv.org/abs/2406.18518} {Apigen: Automated pipeline
  for generating verifiable and diverse function-calling datasets}.
\newblock \emph{arXiv preprint arXiv:2406.18518}.

\bibitem[{Lu et~al.(2025)Lu, Holleis, Zhang, Aumayer, Nan, Bai, Ma, Ma, Li,
  Yin, Wang, and Pang}]{lu2025toolsandbox}
Jiarui Lu, Thomas Holleis, Yizhe Zhang, Bernhard Aumayer, Feng Nan, Felix Bai,
  Shuang Ma, Shen Ma, Mengyu Li, Guoli Yin, Zirui Wang, and Ruoming Pang. 2025.
\newblock Toolsandbox: A stateful, conversational, interactive evaluation
  benchmark for llm tool use capabilities.
\newblock In \emph{Findings of the Association for Computational Linguistics:
  NAACL 2025}.

\bibitem[{{MiniMax}(2026)}]{minimax2026m27}
{MiniMax}. 2026.
\newblock \href {https://www.minimax.io/news/minimax-m27-en} {{MiniMax M2.7}:
  Early echoes of self-evolution}.

\bibitem[{OpenAI(2023)}]{openai2023gpt4}
OpenAI. 2023.
\newblock \href {https://arxiv.org/abs/2303.08774} {{GPT-4} technical report}.
\newblock \emph{arXiv preprint arXiv:2303.08774}.

\bibitem[{{OpenAI}(2026)}]{openai2026gpt55}
{OpenAI}. 2026.
\newblock \href {https://deploymentsafety.openai.com/gpt-5-5/gpt-5-5.pdf}
  {{GPT-5.5 System Card}}.

\bibitem[{Ouyang et~al.(2022)Ouyang, Wu, Jiang, Almeida, Wainwright, Mishkin,
  Zhang, Agarwal, Slama, Ray et~al.}]{ouyang2022training}
Long Ouyang, Jeffrey Wu, Xu~Jiang, Diogo Almeida, Carroll Wainwright, Pamela
  Mishkin, Chong Zhang, Sandhini Agarwal, Katarina Slama, Alex Ray, and 1
  others. 2022.
\newblock \href {https://arxiv.org/abs/2203.02155} {Training language models to
  follow instructions with human feedback}.
\newblock \emph{Advances in Neural Information Processing Systems},
  35:27730--27744.

\bibitem[{Park et~al.(2023)Park, O'Brien, Cai, Morris, Liang, and
  Bernstein}]{park2023generative}
Joon~Sung Park, Joseph~C O'Brien, Carrie~J Cai, Meredith~Ringel Morris, Percy
  Liang, and Michael~S Bernstein. 2023.
\newblock \href {https://arxiv.org/abs/2304.03442} {Generative agents:
  Interactive simulacra of human behavior}.
\newblock In \emph{Proceedings of the 36th Annual ACM Symposium on User
  Interface Software and Technology}.

\bibitem[{Patil et~al.(2025)Patil, Mao, Yan, Ji, Suresh, Stoica, and
  Gonzalez}]{patil2025bfcl}
Shishir~G. Patil, Huanzhi Mao, Fanjia Yan, Charlie Cheng-Jie Ji, Vishnu Suresh,
  Ion Stoica, and Joseph~E. Gonzalez. 2025.
\newblock The berkeley function calling leaderboard (bfcl): From tool use to
  agentic evaluation of large language models.
\newblock In \emph{Proceedings of the 42nd International Conference on Machine
  Learning}, pages 48371--48392.

\bibitem[{{Qwen Team}(2025)}]{qwen2025qwen3}
{Qwen Team}. 2025.
\newblock \href {https://arxiv.org/abs/2505.09388} {Qwen3 technical report}.
\newblock \emph{arXiv preprint arXiv:2505.09388}.

\bibitem[{Schick et~al.(2023)Schick, Dwivedi-Yu, Dess{\`i}, Raileanu, Lomeli,
  Hambro, Zettlemoyer, Cancedda, and Scialom}]{schick2023toolformer}
Timo Schick, Jane Dwivedi-Yu, Roberto Dess{\`i}, Roberta Raileanu, Maria
  Lomeli, Eric Hambro, Luke Zettlemoyer, Nicola Cancedda, and Thomas Scialom.
  2023.
\newblock \href {https://arxiv.org/abs/2302.04761} {Toolformer: Language models
  can teach themselves to use tools}.
\newblock \emph{Advances in Neural Information Processing Systems}, 36.

\bibitem[{Shaikh et~al.(2025)Shaikh, Sapkota, Rizvi, Horvitz, Park, Yang, and
  Bernstein}]{shaikh2025gum}
Omar Shaikh, Shardul Sapkota, Shan Rizvi, Eric Horvitz, Joon~Sung Park, Diyi
  Yang, and Michael~S. Bernstein. 2025.
\newblock \href {https://doi.org/10.1145/3746059.3747722} {Creating general
  user models from computer use}.
\newblock In \emph{Proceedings of the 38th Annual ACM Symposium on User
  Interface Software and Technology (UIST)}.

\bibitem[{Tseng et~al.(2024)Tseng, Huang, Hsiao, Chen, Huang, Meng, and
  Chen}]{tseng2024two}
Yu-Min Tseng, Yu-Chao Huang, Teng-Yun Hsiao, Wei-Lin Chen, Chao-Wei Huang,
  Yu~Meng, and Yun-Nung Chen. 2024.
\newblock Two tales of persona in llms: A survey of role-playing and
  personalization.
\newblock In \emph{Findings of the Association for Computational Linguistics:
  EMNLP 2024}, pages 16612--16631.

\bibitem[{Wei et~al.(2025)Wei, Karina, Chen, Perez, Bowman, Kaplan, Shlegeris,
  and Fedus}]{wei2025browsecomp}
Jason Wei, Nguyen Karina, Hyung~Won Chen, Ethan Perez, Samuel~R. Bowman, Jared
  Kaplan, Buck Shlegeris, and Liam Fedus. 2025.
\newblock \href {https://arxiv.org/abs/2504.12516} {{BrowseComp}: A simple yet
  challenging benchmark for browsing agents}.
\newblock \emph{arXiv preprint arXiv:2504.12516}.

\bibitem[{Wu et~al.(2023)Wu, Bansal, Zhang, Wu, Li, Zhu, Jiang, Zhang, Zhang,
  Liu et~al.}]{wu2023autogen}
Qingyun Wu, Gagan Bansal, Jieyu Zhang, Yiran Wu, Beibin Li, Erkang Zhu,
  Li~Jiang, Xiaoyun Zhang, Shaokun Zhang, Jiale Liu, and 1 others. 2023.
\newblock \href {https://arxiv.org/abs/2308.08155} {{AutoGen}: Enabling
  next-gen {LLM} applications via multi-agent conversation}.
\newblock \emph{arXiv preprint arXiv:2308.08155}.

\bibitem[{{Xiaomi AI Lab}(2026)}]{xiaomi2026mimov2pro}
{Xiaomi AI Lab}. 2026.
\newblock \href {https://huggingface.co/XiaomiMiMo/MiMo-V2-Pro} {{MiMo-V2-Pro}:
  Hugging face model repository}.

\bibitem[{{Xiaomi LLM-Core Team}(2026)}]{xiaomi2026mimov2flash}
{Xiaomi LLM-Core Team}. 2026.
\newblock \href {https://arxiv.org/abs/2601.02780} {Mimo-v2-flash technical
  report}.
\newblock \emph{arXiv preprint arXiv:2601.02780}.

\bibitem[{Xie et~al.(2024)Xie, Zhang, Chen, Li, Zhao, Cao, Hua, Cheng, Shin,
  Lei et~al.}]{xie2024osworld}
Tianbao Xie, Danyang Zhang, Jixuan Chen, Xiaochuan Li, Siheng Zhao, Ruisheng
  Cao, Tainuo Hua, Zhoujun Cheng, Dongchan Shin, Fangyu Lei, and 1 others.
  2024.
\newblock \href {https://arxiv.org/abs/2404.07972} {{OSWorld}: Benchmarking
  multimodal agents for open-ended tasks in real computer environments}.
\newblock \emph{arXiv preprint arXiv:2404.07972}.

\bibitem[{Yao et~al.(2024)Yao, Shinn, Razavi, and Narasimhan}]{yao2024taubench}
Shunyu Yao, Noah Shinn, Parth Razavi, and Karthik Narasimhan. 2024.
\newblock \href {https://arxiv.org/abs/2406.12045} {$\tau$-bench: A benchmark
  for tool-agent-user interaction in real-world domains}.
\newblock \emph{arXiv preprint arXiv:2406.12045}.

\bibitem[{Yao et~al.(2023)Yao, Zhao, Yu, Du, Shafran, Narasimhan, and
  Cao}]{yao2023react}
Shunyu Yao, Jeffrey Zhao, Dian Yu, Nan Du, Izhak Shafran, Karthik Narasimhan,
  and Yuan Cao. 2023.
\newblock \href {https://arxiv.org/abs/2210.03629} {{ReAct}: Synergizing
  reasoning and acting in language models}.
\newblock \emph{International Conference on Learning Representations}.

\bibitem[{Ye and collaborators(2026)}]{ye2026claweval}
Bowen Ye and collaborators. 2026.
\newblock \href {https://arxiv.org/abs/2604.06132} {Claw-eval: End-to-end
  transparent benchmark for real-world agentic systems}.
\newblock \emph{Preprint}, arXiv:2604.06132.

\bibitem[{Yu et~al.(2026)Yu, Liu, Yang, Li, Zhang, Feng, and
  Zhang}]{yu2026wildtoolbench}
Peijie Yu, Wei Liu, Yifan Yang, Jinjian Li, Zelong Zhang, Xiao Feng, and Feng
  Zhang. 2026.
\newblock Benchmarking llm tool-use in the wild.
\newblock In \emph{International Conference on Learning Representations}.

\bibitem[{Zhao et~al.(2024)Zhao, Ren, Hessel, Cardie, Choi, and
  Deng}]{zhao2024wildchat}
Wenting Zhao, Xiang Ren, Jack Hessel, Claire Cardie, Yejin Choi, and Yuntian
  Deng. 2024.
\newblock \href {https://arxiv.org/abs/2405.01470} {Wildchat: 1m chatgpt
  interaction logs in the wild}.
\newblock \emph{arXiv preprint arXiv:2405.01470}.

\bibitem[{Zheng et~al.(2024)Zheng, Chiang, Sheng, Zhuang, Wu, Zhuang, Lin, Li,
  Li, Xing et~al.}]{zheng2024lmsyschat}
Lianmin Zheng, Wei-Lin Chiang, Ying Sheng, Siyuan Zhuang, Zhanghao Wu, Yonghao
  Zhuang, Zi~Lin, Zhuohan Li, Dacheng Li, Eric~P Xing, and 1 others. 2024.
\newblock \href {https://arxiv.org/abs/2309.11998} {Lmsys-chat-1m: A
  large-scale real-world llm conversation dataset}.
\newblock \emph{International Conference on Learning Representations}.

\bibitem[{Zhou et~al.(2024)Zhou, Xu, Zhu, Zhou, Lo, Sridhar, Cheng, Bisk,
  Fried, Alon et~al.}]{zhou2023webarena}
Shuyan Zhou, Frank~F. Xu, Hao Zhu, Xuhui Zhou, Robert Lo, Abishek Sridhar,
  Xianyi Cheng, Yonatan Bisk, Daniel Fried, Uri Alon, and 1 others. 2024.
\newblock \href {https://arxiv.org/abs/2307.13854} {{WebArena}: A realistic web
  environment for building autonomous agents}.
\newblock In \emph{International Conference on Learning Representations}.

\end{thebibliography}

\appendix

\section{Real-Session Analysis Details}
\label{sec:analysis_details}

We use deterministic regular-expression matching to derive the descriptive statistics in Section~\ref{sec:analysis}. Before matching, user messages are normalized by lowercasing English text, converting full-width punctuation to half-width punctuation, and stripping URLs and code blocks. Task categories in Figure~\ref{fig:task_type_multiturn} are assigned from the first user message using priority-ordered regex families. Representative English keywords include \texttt{python}, \texttt{pandas}, \texttt{sql}, \texttt{excel}, \texttt{csv}, \texttt{plot}, \texttt{chart}, and \texttt{analy[sz]e} for data analysis; \texttt{docker}, \texttt{kubernetes}, \texttt{linux}, \texttt{server}, \texttt{deploy}, \texttt{nginx}, \texttt{ssh}, \texttt{cron}, and \texttt{devops} for system/DevOps; \texttt{code}, \texttt{bug}, \texttt{function}, \texttt{class}, \texttt{api}, \texttt{compile}, \texttt{test}, and \texttt{repo} for coding; \texttt{file}, \texttt{pdf}, \texttt{docx}, \texttt{rename}, \texttt{merge}, and \texttt{convert} for file operations; \texttt{wechat}, \texttt{slack}, \texttt{discord}, \texttt{bot}, \texttt{message}, and \texttt{notification} for IM/bot interactions; \texttt{write}, \texttt{rewrite}, \texttt{draft}, \texttt{email}, \texttt{summary}, and \texttt{polish} for writing; and \texttt{search}, \texttt{find}, \texttt{lookup}, \texttt{what is}, and \texttt{who is} for information retrieval. For Chinese sessions, we use translated counterparts of the same task and correction lexicons. If multiple categories match, we use the first category in this priority order; sessions with no category match are retained in aggregate statistics but excluded from the category plot.

We identify explicit user correction signals by matching follow-up user messages against correction-oriented expressions, including \texttt{wrong}, \texttt{incorrect}, \texttt{not right}, \texttt{mistake}, \texttt{error}, \texttt{failed}, \texttt{doesn't work}, \texttt{try again}, \texttt{fix}, \texttt{redo}, \texttt{you missed}, and \texttt{not what I meant}, plus translated Chinese counterparts maintained in the same lexicon. A session is counted as containing a user correction if any post-initial user message matches this set. We mark a correction as post-tool when the immediately preceding assistant step contains at least one tool call or tool return.

For Figure~\ref{fig:failure_modes}, correction messages are further classified by keyword families. \emph{Context loss} matches references to forgotten or previously stated information, such as \texttt{forgot}, \texttt{earlier}, \texttt{previous}, \texttt{as I said}, \texttt{context}, and \texttt{again}. \emph{Wrong approach} matches method-level objections such as \texttt{wrong approach}, \texttt{not this way}, \texttt{should use}, and \texttt{instead}. \emph{Format errors} matches output-shape requests such as \texttt{format}, \texttt{json}, \texttt{table}, \texttt{markdown}, and \texttt{schema}. \emph{Instruction ignored} matches \texttt{ignore}, \texttt{didn't follow}, \texttt{as requested}, and \texttt{I asked}. \emph{Hallucination} matches fabrication cues such as \texttt{hallucinat}, \texttt{made up}, \texttt{not real}, \texttt{fabricat}, and \texttt{source?}. Chinese messages are matched with translated equivalents of these five keyword families. When multiple failure families match the same correction, we count all matched families for the failure-mode histogram; the session-level correction rate remains binary.

\section{Framework Design Supplements}

\subsection{Interaction Structure}

Figure~\ref{fig:interaction_structure} details the information flow between the PersonaForge user simulator and the target agentic system. The simulator's hidden context (SOUL prompt, connected memories, and task scenario) is never visible to the target system, which sees only the accumulated user messages and its own tool outputs.

\subsection{Persona Coherence}

\begin{figure}[t]
    \centering
    \includegraphics[width=\columnwidth]{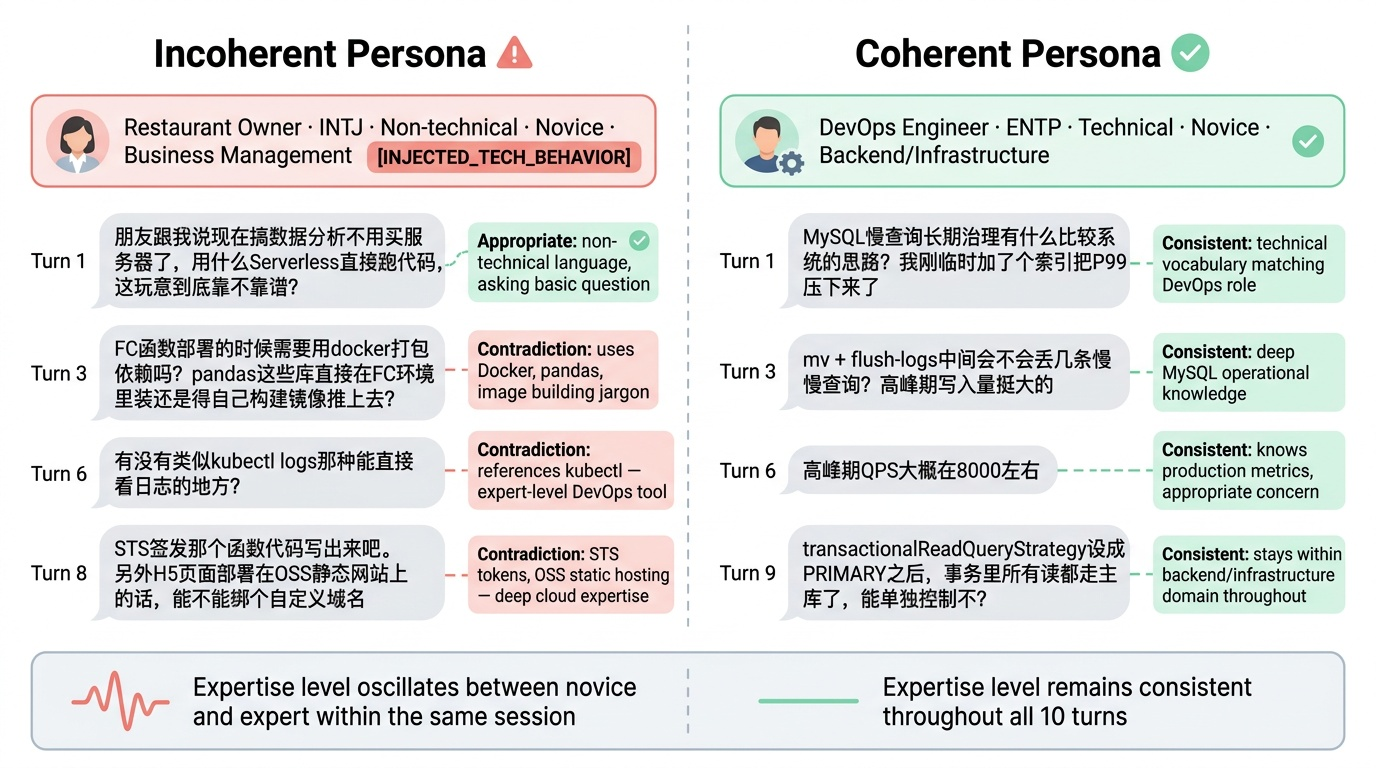}
    \caption{Effect of persona coherence on simulated user behavior. \textbf{Left}: inconsistent persona dimensions produce contradictory behavior within the same session. \textbf{Right}: coherent persona dimensions yield consistent behavior throughout the trajectory.}
    \label{fig:coherence_example}
\end{figure}

Figure~\ref{fig:coherence_example} contrasts two simulated conversations: one with incoherent persona dimensions (a non-technical user paired with expert-level jargon) that produces contradictory behavior, and one with coherent dimensions that yields consistent behavior throughout. This validates the coherence constraints described in Section~\ref{sec:persona}.

\subsection{Profile Construction}
\label{sec:construction_appendix}

\begin{figure*}[!htbp]
    \centering
    \begin{subfigure}[t]{0.49\textwidth}
        \centering
        \includegraphics[width=\textwidth]{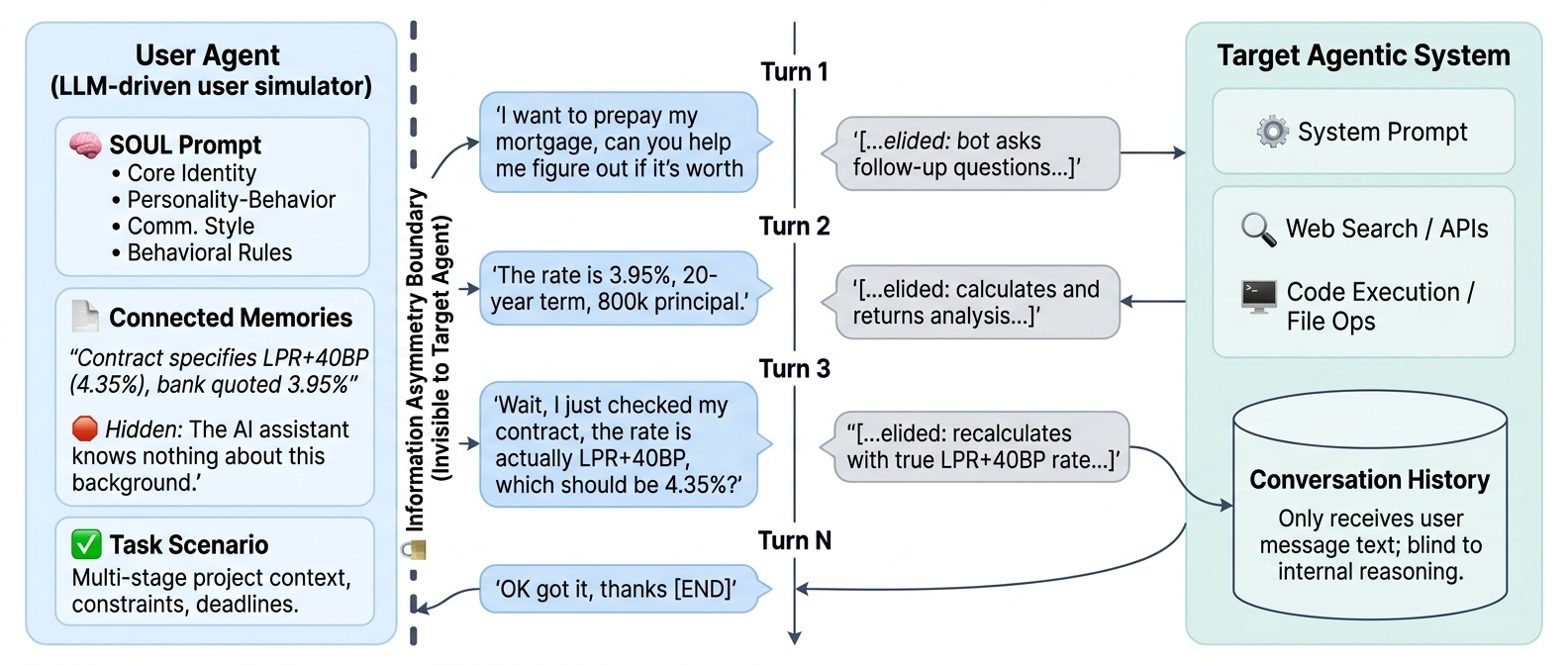}
        \caption{Information asymmetry between the user simulator and the target agentic system.}
        \label{fig:interaction_structure}
    \end{subfigure}\hfill
    \begin{subfigure}[t]{0.49\textwidth}
        \centering
        \includegraphics[width=\textwidth]{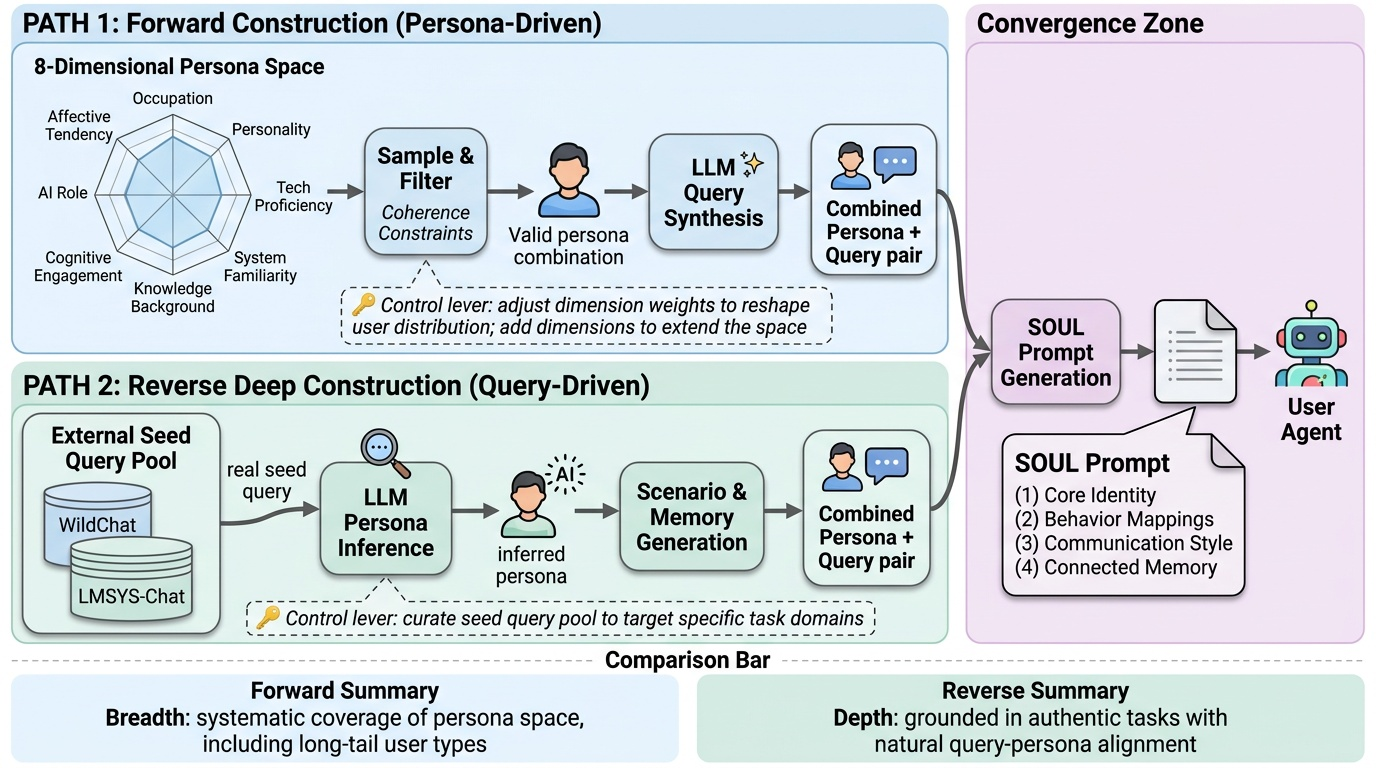}
        \caption{Reverse Deep Construction pipeline.}
        \label{fig:construction_paths}
    \end{subfigure}
    \caption{Framework illustrations: (a) information asymmetry between the user simulator and the target system; (b) reverse construction from real seed queries.}
    \label{fig:framework_figs}
\end{figure*}

Figure~\ref{fig:construction_paths} illustrates the Reverse Deep Construction pipeline described in Section~\ref{sec:construction}: starting from real seed queries, an LLM infers persona profiles and generates task scenarios, converging on the same SOUL prompt structure. We also explored a complementary \emph{forward} (persona-driven) path that samples a persona combination first and then generates a seed query grounded in that persona. While this provides direct control over user-type coverage, we found that Reverse Deep Construction alone produces sufficient diversity when seeded from large-scale real-query corpora, and all PersonaForge synthetic training data used in our experiments is generated via the reverse path.

\noindent\textbf{Forward vs.\ Reverse: a case comparison.} To illustrate why reverse-constructed profiles produce richer simulator state than forward-sampled ones, we present two matched cases. In the finance domain, a forward-sampled profile (a realtor with novice-level finance knowledge) produces memories that are plausible but disjoint: a colleague's Excel template for computing holdings yield, a pharma stock position requiring PE/PB ratios, and a client's down-payment parking for two months. The reverse-constructed profile (a finance manager) converges on a single ongoing scenario---an Ireland SPV acquisition---viewed from three complementary angles: a cross-border tax-structure proposal, a side-by-side comparison of corporate income tax and capital gains tax across Ireland, Singapore, and Hong Kong, and the registration procedure with Irish Revenue. Table~\ref{tab:case_engineering} presents the same comparison for the engineering domain. Each profile is summarized by its persona descriptor and three representative connected memory entries, the dominant component of the SOUL prompt and the single most impactful element identified in our ablation (Appendix~\ref{sec:soul_ablation_tables}). Memories were originally generated in Chinese; we translate them to English here for readability and preserve domain-specific terms.

\begin{table*}[t]
\centering
\small
\setlength{\tabcolsep}{4pt}
\renewcommand{\arraystretch}{1.15}
\begin{tabular}{p{0.10\linewidth}p{0.41\linewidth}p{0.41\linewidth}}
\toprule
& \textbf{Forward (persona-driven)} & \textbf{Reverse Deep Construction} \\
\midrule
Persona & Product Manager $\cdot$ ISFP $\cdot$ Semi-tech $\cdot$ Heavy user $\cdot$ Frontend / Product & Software Engineer $\cdot$ INTJ $\cdot$ Tech-expert $\cdot$ Heavy user $\cdot$ AI Agent Builder \\
\midrule
Memory 1 & Tomorrow I sync with the designer on the user-settings page interaction. I have a rough left-tab / right-content layout in mind, but I haven't worked out per-module fields and states. & Just hit a tool-calling JSON instability problem in the agent pipeline---occasional extra fields and nested misalignment crash downstream parsing. I need to add a hard prompt-level constraint and want a structured-output design quickly. \\
Memory 2 & Frontend wants to switch state management from \emph{Redux to Zustand} for ``lighter weight.'' I know roughly what Redux does but only know Zustand by name. & I've been patching structured output with regex post-processing---too brittle. I should systematically look at \emph{constrained decoding} and design a general layer that fits the existing architecture. \\
Memory 3 & A new feature launches soon and needs a first-time onboarding paragraph; my draft reads too dry, so I want different tones to choose from. & Next week's architecture review will discuss output-format constraints for agents; I need a horizontal comparison of structured-output tools---\emph{constraint granularity, compatibility with mainstream inference frameworks, streaming support}---no concept primers. \\
\bottomrule
\end{tabular}
\caption{Profile comparison in the engineering domain. The reverse-constructed profile is grounded in a specific technical pain point (LLM agent output reliability) with concrete domain vocabulary; the forward-sampled profile's memories are plausible but unrelated peripheral tasks of a product manager.}
\label{tab:case_engineering}
\end{table*}

\noindent\textbf{Observations.} Three differences are consistent across both pairs and explain why the reverse path supplies a stronger training signal. (1)~\emph{Anchor specificity}: reverse-constructed memories carry real-world proper nouns and quantitative references (\emph{Irish Revenue, CGT, Ireland SPV, constrained decoding, streaming}) that an LLM is unlikely to fabricate without grounding, whereas forward-sampled memories rely on generic concepts (PE/PB ratios, Redux, money-market funds). (2)~\emph{Connected-memory depth}: reverse memories converge on a single ongoing scenario viewed from complementary angles---an Ireland SPV acquisition seen as tax planning / cross-jurisdiction comparison / Revenue filing procedure---giving the simulator latent context to reveal across turns; forward memories are plausible but disjoint incidents that leave no hidden context to disclose progressively. (3)~\emph{User--task coherence}: reverse profiles tie the persona's professional identity to the same authentic task that produced the seed query (a finance manager facing an Ireland CGT question, an engineer with a real agent-pipeline output-format problem); forward sampling occasionally pairs a tangential occupation with the knowledge background (a real-estate agent issuing equity-valuation queries), which the simulator must paper over at generation time. These observations are consistent with the SOUL ablation in Appendix~\ref{sec:soul_ablation_tables}, where connected memory ranks as the single most impactful component for downstream task completion.

\subsection{Persona Dimension Distributions}
\label{sec:persona_detail}

\begin{figure*}[!htbp]
    \centering
    \includegraphics[width=\textwidth]{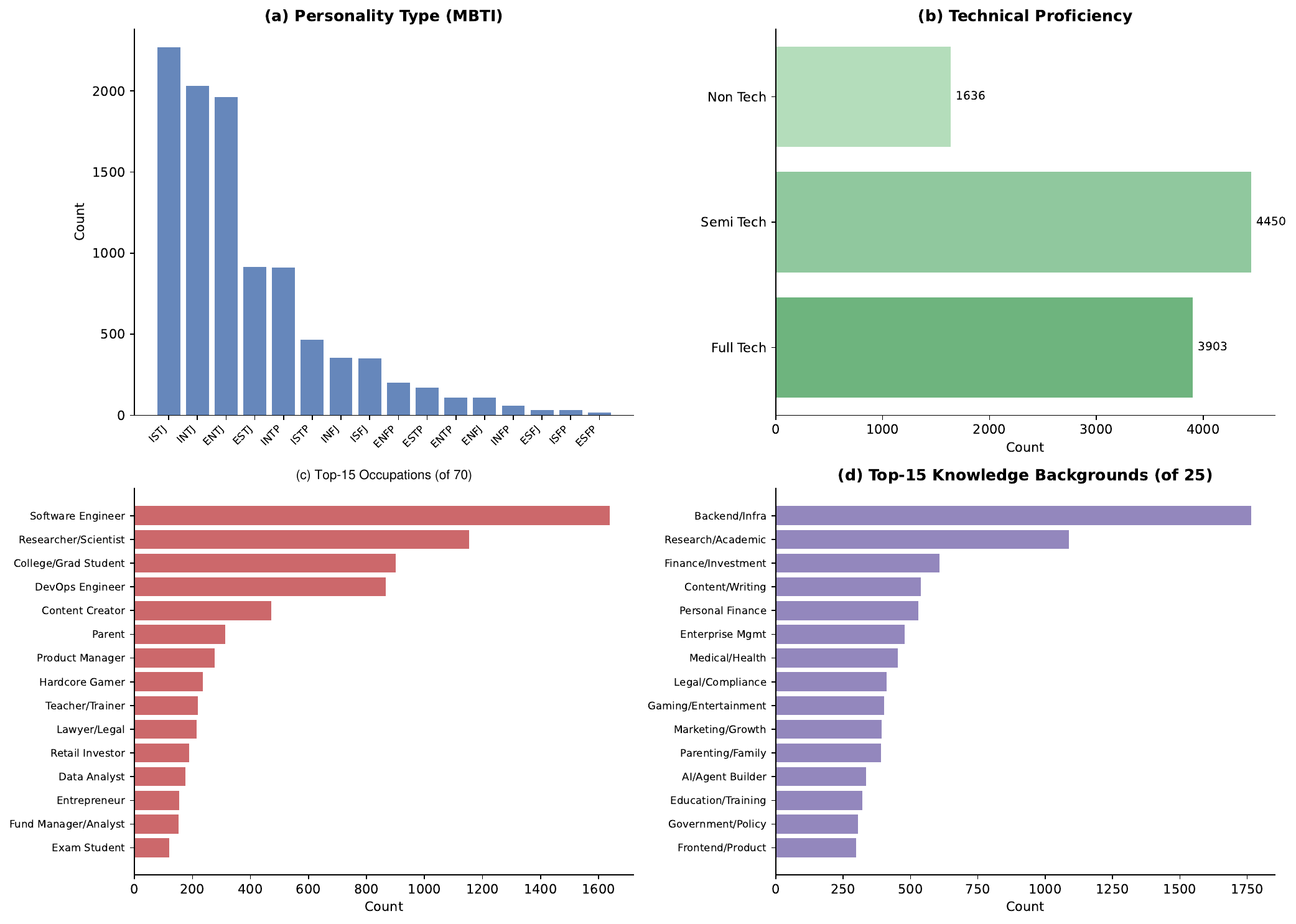}
    \vspace{-4pt}
    \caption{Distribution of persona dimensions. (a)~MBTI types (16 types). (b)~Technical proficiency (3 levels). (c)~Top-15 occupations (of 70). (d)~Top-15 knowledge backgrounds (of 25).}
    \label{fig:persona_dist}
\end{figure*}

Figure~\ref{fig:persona_dist} shows the detailed distribution of each persona dimension in the synthesized dataset. The MBTI distribution is skewed toward introverted judging types (ISTJ, INTJ, ISFJ), which together account for roughly 40\% of all profiles; this reflects the dominance of detail-oriented, task-focused personas in the seed-query pool. Technical proficiency splits into three tiers: semi-technical users form the largest group (44\%), followed by full-technical (37\%) and non-technical (19\%), consistent with the developer-heavy composition of the source corpora. The top-15 occupations cover software engineering, research, content creation, and product management, with a long tail of 55 additional types. Knowledge backgrounds are similarly long-tailed: backend infrastructure and academic research lead, while domains such as parenting, gaming, and legal compliance each contribute fewer than 5\% of profiles.

\subsection{Data Quality Analysis}
\label{sec:data_quality}

Table~\ref{tab:quality_cleaning} reports the effect of each quality control stage on the final dataset. Pre-generation filtering removes 12\% of persona combinations due to coherence violations and 8\% of seed queries for insufficient complexity. Runtime role-reversal detection flags 7\% of sessions. Post-processing removes 4\% of sessions due to degenerate patterns (infinite loops, repetitive exchanges). The listed rates are stage-level diagnostics rather than a strict multiplicative cascade; additional post-processing bookkeeping is not itemized in the table. As a rough estimate, the compounded retention rate across all four stages is approximately 74\%, meaning roughly one in four generated sessions is discarded before entering the training corpus. The two pre-generation filters account for the largest share of removals, underscoring the importance of persona coherence and seed complexity as upstream quality gates.

\begin{table}[H]
  \centering
  \small
  \begin{tabular}{lcc}
    \toprule
    \chg{\textbf{Stage}} & \chg{\textbf{Removed}} & \chg{\textbf{Ret.}} \\
    \midrule
    Persona coherence  & \chg{12\%} & \chg{88\%} \\
    Query complexity    & \chg{8\%}  & \chg{92\%} \\
    Role reversal       & \chg{7\%}  & \chg{93\%} \\
    Degenerate patterns & \chg{4\%}  & \chg{96\%} \\
    \bottomrule
  \end{tabular}
  \caption{Data quality control pipeline statistics.}
  \label{tab:quality_cleaning}
\end{table}

\noindent We note that runtime role reversal is the most subtle filter: the 1{,}500-character threshold was calibrated on real-user data, where 99.2\% of follow-up messages fall below this cutoff. Raising the threshold to 2{,}000 characters drops flagged sessions to 3\% but admits borderline role-reversed turns; lowering to 1{,}000 characters flags 12\% but may discard legitimate long messages (e.g., users pasting code snippets).


\FloatBarrier\clearpage

\section{Rubric Design and Examples}
\label{sec:rubric_examples}

Each task in PersonaForge-Bench is accompanied by a detailed rubric specifying \textbf{MUST\_ASK} clarification items, reference calculations with tolerance windows (e.g., $\pm$100 yuan, $\pm$0.5\%), embedded contradictions or traps, and step-by-step evaluation chains.

\subsection{Benchmark Composition}
\label{sec:benchmark_composition}

PersonaForge-Bench contains 138 tasks across Chinese and English scenarios: 110 tasks are in Chinese and 28 tasks are in English. The language label is specified in each task file and is used only to define the user-facing scenario and rubric language; the same multi-turn evaluation protocol and four scoring dimensions are applied to both subsets.

\subsection{Benchmark Design Principles}

Every task requires sustained multi-turn interaction and is designed around three principles. \emph{Progressive information disclosure}: the user persona reveals details only when the assistant asks the right questions. \emph{Embedded contradictions}: deliberate inconsistencies test whether the assistant can detect and reconcile conflicting information. \emph{Domain-specific rubrics}: each task specifies required clarification items, reference calculations with tolerance windows, and step-by-step decision chains.

\subsection{Scoring Dimensions}

Each task is scored along four dimensions, each weighted equally at 25\%, motivated by the real-user interaction patterns identified in Section~\ref{sec:analysis}. \textbf{Interaction Efficiency} measures how many conversation turns and tool calls are required to reach a satisfactory answer; fewer turns with comparable quality score higher. \textbf{Tool Appropriateness} evaluates whether the assistant selects the correct tools for each sub-task (programmatic computation via Bash, web search for current information, web\_fetch for verification, file operations for document manipulation), whether search queries are well-crafted, and whether tool calls succeed on the first attempt. \textbf{Task Completion} assesses whether the assistant delivers a final, actionable answer covering all user requirements, discovers hidden needs through follow-up questions, and detects embedded contradictions in the user's information. \textbf{Response Quality} evaluates structural clarity (tables, headings, bullet points), factual and numerical accuracy, and directness, specifically whether the response gets to the point without unnecessary preamble.

\subsection{Task-Specific Scoring Protocol}

Scoring is \emph{task-specific}: each of the 138 tasks includes a hand-crafted \texttt{judge\_rubric} defining domain-appropriate evaluation criteria and a \texttt{reference\_solution} specifying expected clarification items, correct calculations with tolerance windows, and step-by-step decision chains that a competent assistant should follow. When scoring a conversation, the LLM judge receives (1)~the task's original user prompt, (2)~its reference solution as ground truth, (3)~its domain-specific evaluation criteria, and (4)~the four generic dimension rubrics. This design ensures that the judge evaluates each conversation against what a correct answer \emph{for that specific task} should look like, rather than applying a one-size-fits-all rubric that cannot account for domain-specific requirements such as detecting a 4.35\% vs.\ 3.95\% interest rate contradiction or verifying drug interaction risks for a renally impaired patient.

\subsection{Design Rationale}

The four scoring dimensions are directly motivated by the real-world patterns in Section~\ref{sec:analysis}.
\textbf{Interaction Efficiency} penalizes inefficient tool use in multi-turn sessions, where models average 99.3 tool calls and 36.0\% of correction signals occur after tool execution.
\textbf{Tool Appropriateness} targets the wrong-approach errors (2.1K occurrences) and instruction-ignored failures (714) identified in our failure analysis, rewarding correct first-attempt tool selection.
\textbf{Task Completion} reflects the inherent incompleteness of real queries; users progressively disclose requirements, so the assistant must discover hidden needs through follow-up.
\textbf{Response Quality} addresses the mismatch between terse user commands (63\% imperative, median 82 characters) and verbose model outputs, rewarding directness and structural clarity.
By weighting efficiency and tool precision at 50\% combined, the benchmark aligns with the terse, correction-heavy patterns of real usage rather than the verbose style of synthetic data.

\subsection{Task-Specific Rubric Examples}

We provide three representative task rubrics to illustrate how the four general dimensions are instantiated with task-specific criteria and weights.

\noindent\textbf{Example 1: Mortgage Prepayment Analysis (C01zh).} The user seeks financial advice on prepaying a mortgage. The task contains an embedded contradiction: the user claims a 3.95\% interest rate, but the contract specifies LPR+40BP (actual rate: 4.35\%). The assistant must collect financial details, detect the discrepancy, compute two repayment strategies (shortening term vs.\ reducing monthly payment), and recommend an approach matched to the user's cash flow.

\noindent\textbf{Example 2: Design Spec Documentation Tools (C074en).} The user needs actionable tool recommendations for auto-generating design specification documentation from Figma files (interaction behaviors, state descriptions, and edge cases). The assistant must balance research thoroughness against practical conciseness: the user needs curated top picks with setup instructions, not an exhaustive academic survey.

\begin{table*}[t]
  \centering
  \small
  \setlength{\tabcolsep}{4pt}
  \begin{tabular}{lp{0.58\linewidth}c}
    \toprule
    \textbf{Dimension} & \textbf{Key Criteria (C01zh)} & \textbf{Wt.} \\
    \midrule
    Interaction Efficiency & $\leq$ 12 turns, $\leq$ 25 tool calls. Penalizes repeated recalculations or searching for general mortgage knowledge. & 0.25 \\
    Tool Appropriateness  & Bash (Python) for amortization computation; web\_search only to verify bank-specific penalty policies. No manual arithmetic. & 0.25 \\
    Task Completion       & Covers both repayment strategies with total interest saved. Provides a concrete, personalized recommendation. & 0.25 \\
    Response Quality      & Clean comparison table. Correctly identifies 4.35\% vs.\ 3.95\% contradiction. No preamble. Proactively flags LPR discrepancy. & 0.25 \\
    \midrule
    \textbf{Dimension} & \textbf{Key Criteria (C074en)} & \textbf{Wt.} \\
    \midrule
    Interaction Efficiency & $\leq$ 12 turns, $\leq$ 15 tool calls. Uses search $\to$ fetch to verify a focused set of tools rather than exhaustively enumerating every option. & 0.25 \\
    Tool Appropriateness  & web\_search to discover candidate tools; web\_fetch on official docs/GitHub to verify maintenance status and features. & 0.25 \\
    Task Completion       & Delivers top 3--5 picks with justification, setup instructions, and a reusable per-component spec template. & 0.25 \\
    Response Quality      & Scannable output: comparison table, clear headings, bullet takeaways. Factual claims are source-verified. Direct and concise. & 0.25 \\
    \bottomrule
  \end{tabular}
  \caption{Task-specific rubric examples. \textbf{Top}: C01zh (mortgage prepayment analysis). \textbf{Bottom}: C074en (design spec documentation tools).}
  \label{tab:rubric_c01}
  \label{tab:rubric_c074}
\end{table*}

\noindent\textbf{Example 3: Container Orchestration Architecture (C11zh).} A DevOps engineer plans a microservice migration to Kubernetes. The assistant must assess the user's infrastructure scale, avoid over-engineering (e.g., not every team needs a full service mesh), and deliver a phased rollout plan.

\noindent\textbf{Rubric summary.} Interaction Efficiency (25\%) penalizes generic ``here are all the CNCF landscape tools'' responses in favor of right-sized recommendations. Tool Appropriateness (25\%) rewards Bash-based resource estimation and targeted web\_search for tool compatibility. Task Completion (25\%) requires concrete migration steps with per-phase checkpoints. Response Quality (25\%) expects architecture diagrams or structured concept maps where helpful, with phased plans and clear completion criteria per phase.

\section{\chg{Comparison with Related Benchmarks}}
\label{sec:benchmark_comparison}

\chg{Table~\ref{tab:benchmark_comparison} situates PersonaForge-Bench against established tool-use and agentic-interaction benchmarks along nine construction axes: task count, languages, whether a live model drives the user, persona diversity, progressive information disclosure, embedded contradictions, corrective-user behavior, human verification, and the style of per-task scoring rubric. PersonaForge-Bench is the only entry that combines a persona-conditioned live user simulator (with hidden context) with progressive disclosure, embedded contradictions, corrective behavior, human-verified tasks, and per-task rubrics.}

\begin{table*}[!htbp]
  \centering
  \scriptsize
  \setlength{\tabcolsep}{2.6pt}
  \resizebox{\textwidth}{!}{%
  \begin{tabular}{lccccccccc}
    \toprule
    \chg{\textbf{Benchmark}} & \chg{\textbf{\#Tasks}} & \chg{\textbf{Lang}} & \chg{\textbf{User sim.}} & \chg{\textbf{Persona div.}} & \chg{\textbf{Prog.\ discl.}} & \chg{\textbf{Contradict.}} & \chg{\textbf{Corrective}} & \chg{\textbf{Human-ver.}} & \chg{\textbf{Rubrics}} \\
    \midrule
    $\tau$-bench / $\tau^2$-bench~\citep{yao2024taubench} & \chg{3 domains} & \chg{en} & \chg{$\checkmark$} & \chg{script-fixed} & \chg{partial} & \chg{$\times$} & \chg{partial} & \chg{$\checkmark$} & \chg{policy} \\
    ToolSandbox~\citep{lu2025toolsandbox} & \chg{1,032} & \chg{en} & \chg{$\checkmark$} & \chg{$\times$} & \chg{partial} & \chg{$\times$} & \chg{$\times$} & \chg{$\checkmark$} & \chg{milestone} \\
    BFCL v3 multi-turn~\citep{patil2025bfcl} & \chg{800} & \chg{en} & \chg{scripted} & \chg{$\times$} & \chg{partial} & \chg{$\times$} & \chg{$\times$} & \chg{$\checkmark$} & \chg{state} \\
    ACEBench (agent split)~\citep{hao2025acebench} & \chg{2,000} & \chg{zh+en} & \chg{$\checkmark$} & \chg{$\times$} & \chg{partial} & \chg{$\times$} & \chg{$\times$} & \chg{$\checkmark$} & \chg{rule} \\
    WildToolBench~\citep{yu2026wildtoolbench} & \chg{1,024} & \chg{en} & \chg{$\times$ (static)} & \chg{$\times$} & \chg{$\checkmark$} & \chg{$\times$} & \chg{$\checkmark$} & \chg{$\checkmark$} & \chg{$\checkmark$} \\
    \midrule
    \textbf{PersonaForge-Bench} & \chg{138} & \chg{zh+en} & \chg{$\checkmark$ (persona + hidden ctx.)} & \chg{$\checkmark$ (4-dim)} & \chg{$\checkmark$} & \chg{$\checkmark$} & \chg{$\checkmark$} & \chg{$\checkmark$} & \chg{$\checkmark$ (per-task)} \\
    \bottomrule
  \end{tabular}%
  }
  \caption{Comparison of agentic-interaction benchmarks across nine construction axes. \emph{User sim.} indicates whether a live model plays the user (BFCL v3 uses scripted multi-turn templates; WildToolBench ships fixed, human-verified dialogues). \emph{Rubrics} marks the dominant scoring mechanism: policy rules ($\tau$-bench), state-based checks (BFCL), milestone evaluation (ToolSandbox), rule-based accuracy (ACEBench), or per-task rubrics (PersonaForge-Bench).}
  \label{tab:benchmark_comparison}
\end{table*}

\section{Judge Reliability and Robustness Analysis}
\label{sec:judge_robustness}

\noindent\textbf{Human validation of judge scores.} To validate the LLM judge's reliability, we conducted a human audit on 60 scored conversations drawn to cover all four dimensions, both models (Base and SFT), and a spread of score levels. Three domain-expert annotators independently scored each conversation using the same dimension-specific rubrics and five-point scale as the judge. The inter-annotator agreement was $\kappa_{hh} = 0.81$ (Cohen's weighted $\kappa$, quadratic weights, averaged over all annotator pairs), providing a human-performance baseline. Judge--human agreement, measured by the weighted Cohen's $\kappa$ between the judge and the averaged human ratings, was $\kappa_{jh} = 0.79$, closely approaching the inter-annotator baseline. The Pearson correlation between judge and averaged human composite scores was $r = 0.84$ ($p < 0.001$). Both metrics confirm the LLM judge as a reliable proxy for human quality assessment.

\noindent\textbf{Scoring protocol robustness.} To verify that our findings are not an artifact of the task-specific scoring protocol, we also evaluated all models using a generic four-dimension rubric without task-specific reference solutions or domain criteria. Both protocols agree on the direction and magnitude of improvement (generic $\Delta = +13.0\%$, task-specific $\Delta = +15.7\%$), with a Pearson correlation of $r = 0.94$ between the two scoring methods.

\noindent\textbf{Cross-judge validation.} To further verify that our scoring results are not an artifact of the specific judge model, we re-scored all 138 Flash tasks across 3 trials (both Base and SFT) using Gemini 3.1 Pro as an independent second judge with identical prompts and rubrics. The two judges exhibit strong agreement on composite scores (Pearson $r = 0.84$, $n = 828$ scored traces). Critically, both independently estimate the PersonaForge gain at nearly identical values: $+15.7\%$ (DeepSeek-V4-Pro) vs.\ $+15.6\%$ (Gemini 3.1 Pro), with a per-task delta correlation of $r = 0.70$ after averaging over 3 trials. These results confirm that the observed improvement is robust to the choice of judge model.

Building on this validation, we further verify that the observed gains are not an artifact of response length or formatting bias, a concern given that SFT increases assistant output by 57.0\%. We conduct three additional robustness checks.

\noindent\textbf{Length-stratified analysis.} Table~\ref{tab:length_stratified} partitions all scored responses from the 3 evaluation trials into four length quartiles and computes the SFT $-$ Base composite difference within each bin. The sample distribution across bins reflects the behavioral difference between models: Base traces are concentrated in shorter bins (144 in Q1 vs.\ 63 for SFT), while SFT traces predominate in longer bins as it produces more substantive responses (165 in Q4 vs.\ 42 for Base). Contrary to the length-bias hypothesis, the SFT advantage is \emph{largest} in the shortest quartile ($+24.8\%$, Q1) and remains positive in every quartile. If the judge merely rewarded longer responses, the gain would be concentrated in longer bins rather than peaking in Q1.

We further verify that the Q1 gain is not an artifact of incomplete or ceiling-limited short traces. After excluding traces with zero assistant output or turn counts at the 25-turn ceiling, the clean Q1 subset (45 Base, 39 SFT traces) yields an even larger SFT advantage of $+29.6\%$ (Base 48.1\% vs.\ SFT 77.7\%), confirming that the Q1 gain reflects genuine quality differences rather than comparing incomplete traces against successful ones.

\begin{table}[H]
  \centering
  \small
  \setlength{\tabcolsep}{4pt}
  \resizebox{\columnwidth}{!}{%
  \begin{tabular}{lccccc}
    \toprule
    \textbf{Length Bin} & \textbf{\#Base} & \textbf{\#SFT} & \textbf{Base} & \textbf{SFT} & $\mathbf{\Delta}$ \\
    \midrule
    Q1 ($<$7.7K chars) & 144 & 63 & 37.1\% & 61.8\% & $+24.8\%$ \\
    Q2 (7.7--13.9K)    & 117 & 90 & 71.8\% & 80.0\% & $+8.2\%$ \\
    Q3 (13.9--22.8K)   & 111 & 96 & 76.6\% & 83.8\% & $+7.2\%$ \\
    Q4 ($>$22.8K chars) & 42 & 165 & 65.8\% & 74.9\% & $+9.1\%$ \\
    \midrule
    All                 & 414 & 414 & 60.4\% & 76.1\% & $+15.7\%$ \\
    \bottomrule
  \end{tabular}%
  }
  \caption{Length-stratified composite scores (138 tasks, 3 scored traces per model per task).}
  \label{tab:length_stratified}
\end{table}

\noindent\textbf{Regression analysis.} We fit a linear model $\text{composite} = \beta_0 + \beta_1 \log(\text{length}) + \beta_2 \cdot \text{is\_SFT}$ over all 828 scored traces (adjusted $R^2 = 0.33$). The length coefficient is $\beta_1 = +0.056$ ($p < 0.001$, SE = 0.006), indicating that each $e$-fold increase in response length corresponds to a 0.056-point score increase. The SFT identity effect is $\beta_2 = +0.087$ ($p = 0.001$, SE = 0.026), meaning that, after controlling for response length, SFT traces score 0.087 points higher on average. Length and SFT identity together explain 33\% of the variance; the SFT effect accounts for the majority of the composite improvement (raw $\Delta = +0.157$), with a complementary but smaller contribution from increased response length. Even after adjusting for length, the SFT improvement remains statistically and practically significant.

\noindent\textbf{Structure and formatting bias.} Table~\ref{tab:structure_bias} stratifies scores by two binary surface features: whether the response contains Markdown tables and whether it uses section headings. In every sub-group, the SFT advantage remains positive and similar in magnitude. Notably, the base model already employs tables (78\% of traces) and headings (88\%) at high rates; SFT modestly increases these proportions (93\% and 99\%), but the within-group gains are comparable across feature values, indicating that structural formatting does not explain the composite improvement. The ``No Headings'' sub-group is excluded due to insufficient sample size (SFT $n = 4$).

\begin{table}[H]
  \centering
  \footnotesize
  \setlength{\tabcolsep}{2.5pt}
  \resizebox{\columnwidth}{!}{%
  \begin{tabular}{lcccc}
    \toprule
    \textbf{Feature} & \textbf{Base (\%)} & \textbf{SFT (\%)} & \textbf{Base Overall} & \textbf{SFT Overall} \\
    \midrule
    Has Table      & 78\% & 93\% & 67.3\% & 77.6\% \\
    No Table       & 22\% &  7\% & 36.6\% & 54.4\% \\
    Has Headings   & 88\% & 99\% & 65.8\% & 77.0\% \\
    \bottomrule
  \end{tabular}%
  }
  \caption{Structure-stratified composite scores.}
  \label{tab:structure_bias}
\end{table}

Taken together, these analyses establish that the reported gains are not artifacts of length or formatting bias. The SFT improvement persists within every length quartile (including a clean subset that excludes zero-output and turn-ceiling traces), within every structural sub-group, and after regression adjustment for response length ($\beta_{\text{SFT}} = +0.087$, $p = 0.001$). PersonaForge training produces genuine interaction quality improvements beyond what increased verbosity alone can explain.

\noindent\chg{\textbf{Cross-simulator validation.} A remaining entanglement is that the evaluation-time user simulator, Claude Opus 4.6, also appears as a scored baseline in Table~\ref{tab:leaderboard}. To measure any same-family advantage, we re-ran the benchmark with GPT-5.5~\citep{openai2026gpt55} as the user simulator while keeping the judge unchanged ($n = 135$ tasks). The untrained Claude reference scores 88.2\% under the Claude Opus 4.6 simulator but 79.7\% under GPT-5.5, confirming one same-family effect on absolute difficulty. For the fine-tuned models, Task Completion remains stable under the new simulator, while Response Quality and Tool Appropriateness dip relative to their own base models; the base models themselves score higher under GPT-5.5 than under Claude Opus 4.6. In every case, the simulator shifts absolute difficulty without changing the ordering of systems, and all published comparisons hold the simulator fixed.}

\section{Additional Experimental Results}
\label{sec:additional_results}
\label{sec:per_domain_detail}

\subsection{\chg{Data Recipe and Transfer Analysis}}
\label{sec:data_recipe_transfer}

\noindent\chg{\textbf{Data recipe vs.\ data quantity.} Table~\ref{tab:matched_scale} reports the matched-scale training controls on Qwen3.5-27B, each run under the identical protocol (budgets, epochs, and base model). Two external corpora fail to reproduce the gains: APIGen-MT-5k~\citep{liu2024apigen}, an external multi-turn corpus, yields $-2.7\%$ composite [CI $-6.0,+0.7$], and ToolACE~\citep{liu2024toolace}, an external single-turn function-calling corpus, yields $-5.5\%$ [$-9.4,-1.6$], a significant regression. Both corpora are strong within their own regime of rigid API orchestration, so we read their failure here as a mismatch of interaction regime. A third control replaces only the construction direction: forward-sampled profiles at equal scale and identical quality control gain nothing significant ($+1.2\%$, n.s.) and degrade Tool Appropriateness ($-5.8\%$, $p<0.001$), confirming that the training signal comes from Reverse Deep Construction itself and cannot be explained by the persona machinery alone. Finally, LoRA ($r{=}32$)~\citep{hu2022lora} on our corpus reproduces the full-SFT profile ($+3.8\%$ [$+1.4,+6.3]$, $p=.005$, versus $+4.1\%$ for full SFT), so the result is not tied to the fine-tuning method. Together with the replay comparison in Section~\ref{sec:ablations}, these controls show that the gain comes from the data \emph{recipe}, namely persona-grounded reverse construction with live interaction, and cannot be explained by data quantity.}

\begin{table}[h]
    \centering
    \footnotesize
    \resizebox{\columnwidth}{!}{%
    \begin{tabular}{lc}
        \toprule
        \chg{\textbf{Control (Qwen3.5-27B)}} & \chg{\textbf{Overall $\Delta$ vs.\ Base (95\% CI)}} \\
        \midrule
        APIGen-MT-5k (external, multi-turn)  & \chg{$-2.7\%$ [$-6.0,+0.7$]} \\
        ToolACE (external, single-turn)      & \chg{$-5.5\%$ [$-9.4,-1.6$]} \\
        Forward-sampled construction         & \chg{$+1.2\%$ [$-1.6,+4.2$]} \\
        LoRA ($r{=}32$) on our corpus        & \chg{$+3.8\%$ [$+1.4,+6.3$]} \\
        Full SFT on our corpus (paper)       & \chg{$+4.1\%$ [$+1.9,+6.4$]} \\
        \bottomrule
    \end{tabular}%
    }
    \caption{\chg{Matched-scale training controls on Qwen3.5-27B, reporting the composite (Overall) $\Delta$ vs.\ Base with paired bootstrap 95\% CIs.}}
    \label{tab:matched_scale}
\end{table}

\noindent\textbf{Transfer on an independent benchmark.} We additionally evaluate on the fully independent third-party BFCL v3 multi-turn~\citep{patil2025bfcl} (800 tasks, rule-scored, no LLM judge): the LoRA ($r{=}32$) model trained on our corpus improves over the base model from 63.0\% to 65.3\%, with gains in every failure category. Together with the held-out \textsc{Claw-Eval} tasks, this further confirms the generalization of PersonaForge training.

\begin{table}[H]
  \centering
  \small
  \setlength{\tabcolsep}{3pt}
  \resizebox{\columnwidth}{!}{%
  \begin{tabular}{lccccc}
    \toprule
    \textbf{Condition} & \textbf{Eff.} & \textbf{Tool} & \textbf{Compl.} & \textbf{Resp.} & \textbf{Overall} \\
    \midrule
    Base & 53.2\% & 73.5\% & 53.5\% & 61.3\% & 60.4\% \\
    + PersonaForge & 69.8\% & 83.7\% & 75.5\% & 75.4\% & 76.1\% \\
    $\Delta$ & $+16.6\%$ & $+10.1\%$ & $+22.0\%$ & $+14.0\%$ & $+15.7\%$ \\
    \bottomrule
  \end{tabular}%
  }
  \caption{Partial MiMo-V2-Flash ablation on PersonaForge-Bench. We report the Base vs.\ PersonaForge SFT comparison here because the full SOUL and user-simulation ablations were conducted on Qwen3.5-27B.}
  \label{tab:mimo_partial_ablation}
\end{table}

\begin{figure}[H]
    \centering
    \includegraphics[width=\columnwidth]{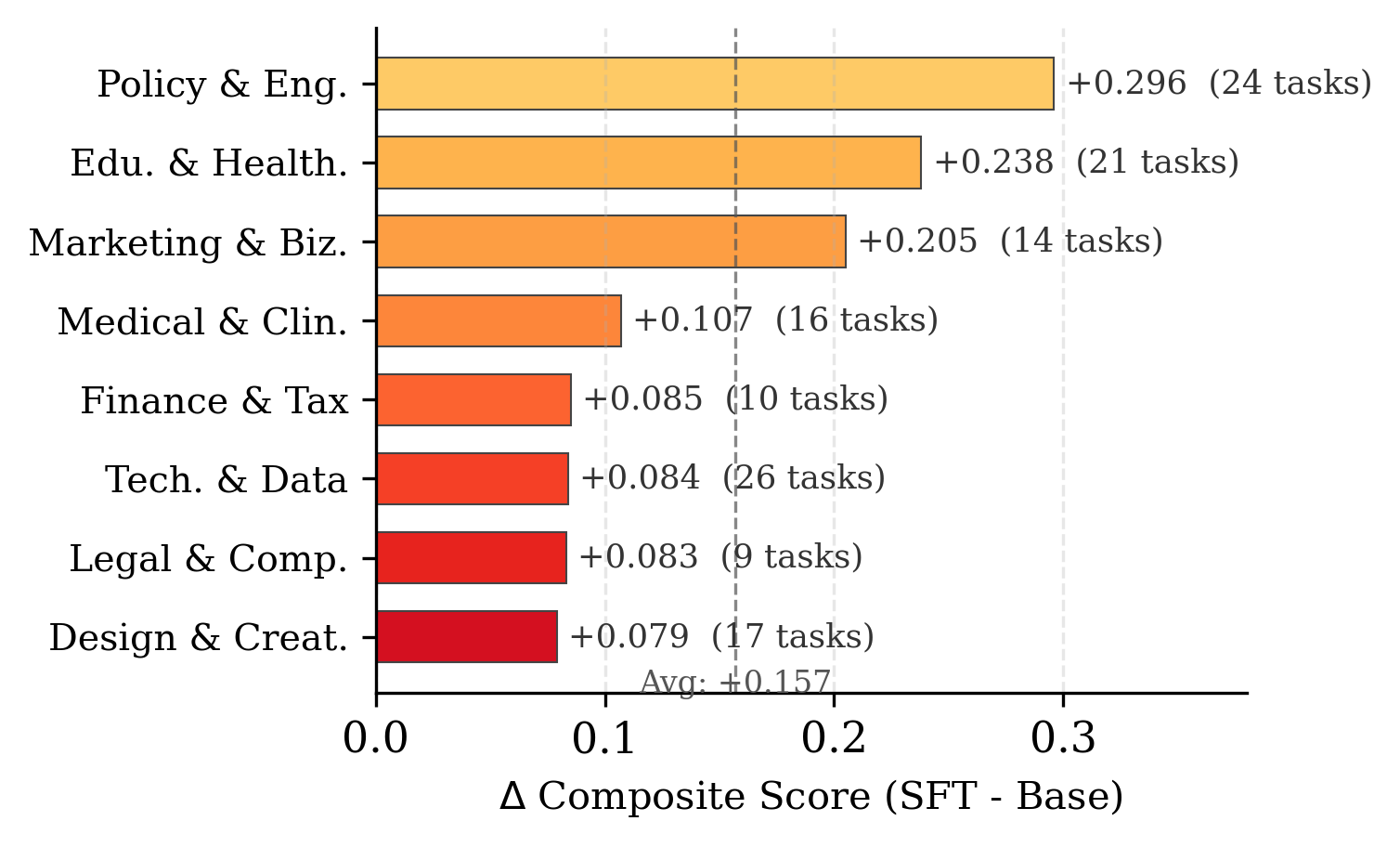}
    \caption{Per-domain Overall scores. $\Delta$ = SFT $-$ Base. Dashed line indicates the average MiMo-V2-Flash gain (+15.7\%).}
    \label{fig:per_domain}
\end{figure}

\begin{figure}[H]
    \centering
    \includegraphics[width=\columnwidth]{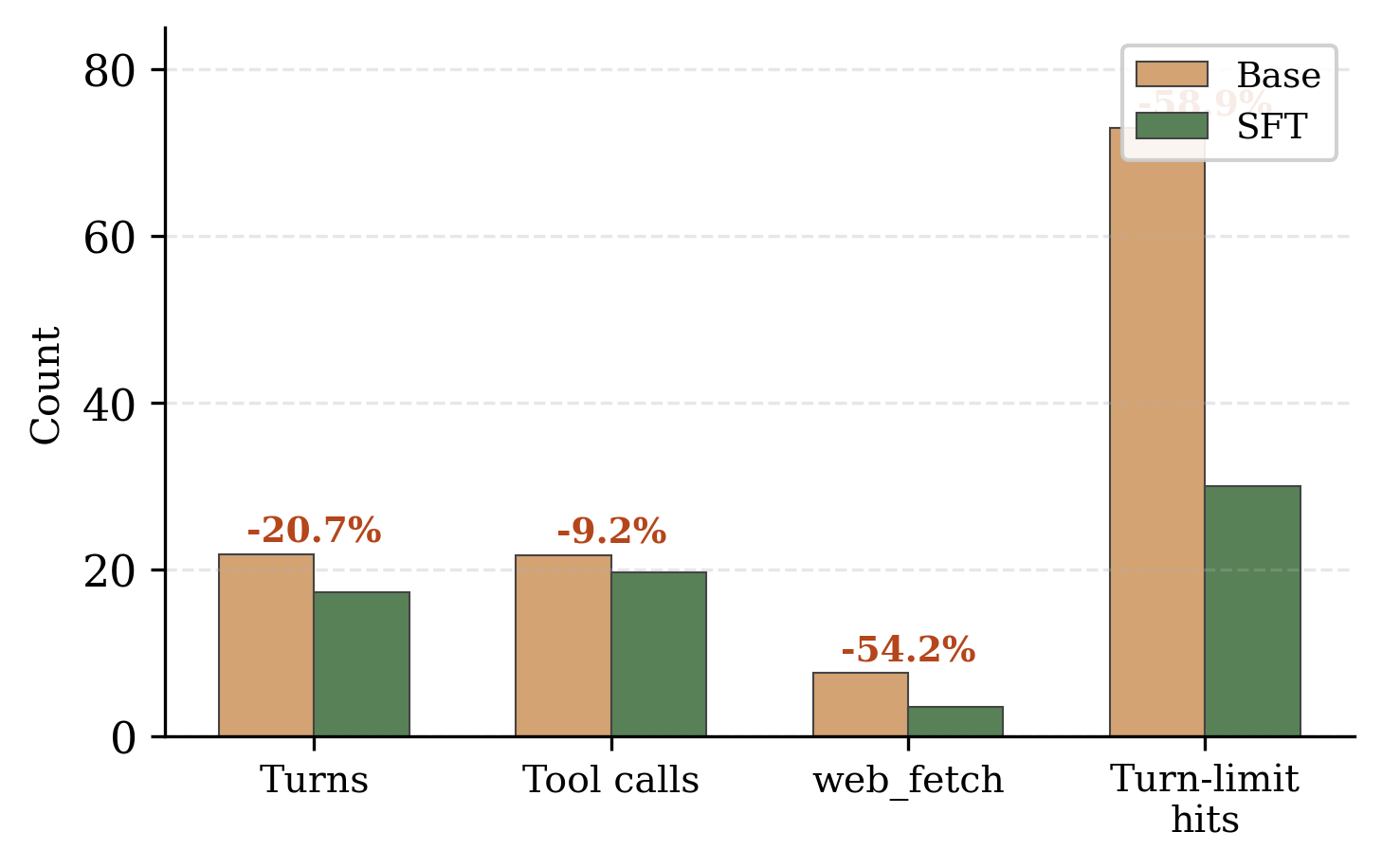}
    \caption{Interaction efficiency comparison. Base (brown) vs.\ SFT (green), with percentage reduction labeled.}
    \label{fig:efficiency}
\end{figure}

\begin{table*}[!htbp]
  \centering
  \scriptsize
  \setlength{\tabcolsep}{2.2pt}
  \begin{tabular}{lccccc|ccccc|c}
    \toprule
    & \multicolumn{5}{c|}{\textbf{Base (MiMo-V2-Flash)}} & \multicolumn{5}{c|}{\textbf{PersonaForge SFT}} & \\
    \cmidrule(lr){2-6} \cmidrule(lr){7-11}
    \textbf{Domain} & \textbf{Eff} & \textbf{Tool} & \textbf{Compl} & \textbf{Resp} & \textbf{Overall} & \textbf{Eff} & \textbf{Tool} & \textbf{Compl} & \textbf{Resp} & \textbf{Overall} & $\Delta$ \\
    \midrule
    Design \& Creative      & 62.1 & 78.2 & 50.3 & 72.9 & 65.9 & 63.5 & 75.3 & 77.1 & 79.1 & 73.8 & $+7.9$ \\
    Technology \& Data      & 62.9 & 76.3 & 65.6 & 65.0 & 67.5 & 72.1 & 82.7 & 76.9 & 71.6 & 75.8 & $+8.4$ \\
    Education \& Healthcare & 50.0 & 69.5 & 58.1 & 71.7 & 62.3 & 79.3 & 91.4 & 86.4 & 87.1 & 86.1 & $+23.7$ \\
    Marketing \& Business   & 48.6 & 67.9 & 47.9 & 52.9 & 54.3 & 70.0 & 83.6 & 70.7 & 74.6 & 74.7 & $+20.4$ \\
    Finance \& Tax          & 58.5 & 87.0 & 54.0 & 49.5 & 62.3 & 69.0 & 91.0 & 58.0 & 65.0 & 70.8 & $+8.5$ \\
    Medical \& Clinical     & 61.6 & 83.4 & 61.4 & 65.6 & 68.0 & 72.5 & 83.8 & 86.3 & 72.5 & 78.8 & $+10.7$ \\
    Legal \& Compliance     & 68.3 & 75.2 & 76.7 & 83.0 & 75.8 & 75.6 & 85.6 & 82.2 & 92.9 & 84.1 & $+8.2$ \\
    Policy \& Engineering   & 26.9 & 60.0 & 26.0 & 37.5 & 37.6 & 60.6 & 79.4 & 63.3 & 65.5 & 67.2 & $+29.6$ \\
    \midrule
    \textbf{Total}          & 53.2 & 73.5 & 53.5 & 61.3 & 60.4 & 69.8 & 83.7 & 75.5 & 75.4 & 76.1 & $\mathbf{+15.7}$ \\
    \bottomrule
  \end{tabular}
  \caption{Per-domain performance of MiMo-V2-Flash (\%). All four dimensions plus Overall. $\Delta$=Overall SFT$-$Base.}
  \label{tab:per_domain_mimo}
\end{table*}

\begin{table*}[!htbp]
  \centering
  \scriptsize
  \setlength{\tabcolsep}{2.2pt}
  \begin{tabular}{lccccc|ccccc|c}
    \toprule
    & \multicolumn{5}{c|}{\textbf{Base (Qwen3.5-27B)}} & \multicolumn{5}{c|}{\textbf{PersonaForge SFT}} & \\
    \cmidrule(lr){2-6} \cmidrule(lr){7-11}
    \textbf{Domain} & \textbf{Eff} & \textbf{Tool} & \textbf{Compl} & \textbf{Resp} & \textbf{Overall} & \textbf{Eff} & \textbf{Tool} & \textbf{Compl} & \textbf{Resp} & \textbf{Overall} & $\Delta$ \\
    \midrule
    Design \& Creative      & 78.8 & 89.2 & 74.7 & 79.4 & 80.5 & 78.3 & 87.5 & 80.2 & 81.7 & 81.9 & $+1.4$ \\
    Technology \& Data      & 75.2 & 89.5 & 73.8 & 71.2 & 77.4 & 78.8 & 91.9 & 83.0 & 77.1 & 82.7 & $+5.3$ \\
    Education \& Healthcare & 72.3 & 86.5 & 67.3 & 70.2 & 74.1 & 74.4 & 84.4 & 71.9 & 80.0 & 77.7 & $+3.6$ \\
    Marketing \& Business   & 79.3 & 91.5 & 79.9 & 76.3 & 81.7 & 83.5 & 91.8 & 78.9 & 83.6 & 84.4 & $+2.7$ \\
    Finance \& Tax          & 73.0 & 87.6 & 74.3 & 69.7 & 76.1 & 73.5 & 91.3 & 75.9 & 76.5 & 79.3 & $+3.2$ \\
    Medical \& Clinical     & 74.9 & 86.1 & 68.3 & 72.8 & 75.5 & 70.1 & 90.5 & 75.3 & 77.8 & 78.4 & $+2.9$ \\
    Legal \& Compliance     & 78.0 & 97.2 & 77.2 & 77.6 & 82.5 & 86.2 & 91.2 & 85.6 & 84.8 & 86.9 & $+4.5$ \\
    Policy \& Engineering   & 67.4 & 83.5 & 59.7 & 67.2 & 69.4 & 71.0 & 87.0 & 69.2 & 74.4 & 75.4 & $+6.0$ \\
    \midrule
    \textbf{Total}          & 74.0 & 88.1 & 70.6 & 72.1 & 76.2 & 76.5 & 89.1 & 76.5 & 78.9 & 80.3 & $\mathbf{+4.1}$ \\
    \bottomrule
  \end{tabular}
  \caption{Per-domain performance of Qwen3.5-27B (\%). All four dimensions plus Overall. $\Delta$=Overall SFT$-$Base.}
  \label{tab:per_domain_qwen}
\end{table*}

\noindent\textbf{Domain assignment.} Per-domain results use deterministic keyword matching on task names. We lowercase each task name, match it against the domain keyword lists used by our evaluation script, and assign the task to the first matched reporting domain in a fixed priority order. The keyword families include: Design \& Creative (\texttt{design}, \texttt{creative}, \texttt{ui\_ux}, \texttt{accessibility}, \texttt{illustration}, \texttt{graphic}, \texttt{visual}, \texttt{typography}, \texttt{branding}); Technology \& Data (\texttt{image\_processing}, \texttt{container}, \texttt{distributed}, \texttt{api\_}, \texttt{database}, \texttt{code\_}, \texttt{programming}, \texttt{devops}, \texttt{sop\_}, \texttt{server}, \texttt{deployment}, \texttt{kubernetes}, \texttt{docker}, \texttt{microservice}, \texttt{backend}, \texttt{frontend}, \texttt{algorithm}); Education \& Healthcare (\texttt{academic}, \texttt{education}, \texttt{teaching}, \texttt{curriculum}, \texttt{research\_method}, \texttt{thesis}, \texttt{literature}, \texttt{paper\_}, \texttt{course\_}, \texttt{student}); Marketing \& Business (\texttt{marketing}, \texttt{business}, \texttt{sales}, \texttt{ecommerce}, \texttt{brand\_}, \texttt{advertising}, \texttt{seo\_}, \texttt{social\_media}, \texttt{content\_}, \texttt{real\_estate}); Finance \& Tax (\texttt{mortgage}, \texttt{finance}, \texttt{tax}, \texttt{investment}, \texttt{accounting}, \texttt{portfolio}, \texttt{income\_tax}, \texttt{stock\_}, \texttt{banking}, \texttt{insurance}, \texttt{loan}); Medical \& Clinical (\texttt{clinical}, \texttt{pharmacy}, \texttt{medical}, \texttt{health}, \texttt{mental\_health}, \texttt{diagnostic}, \texttt{patient}, \texttt{drug\_}, \texttt{disease}, \texttt{surgery}); Legal \& Compliance (\texttt{labor\_law}, \texttt{legal}, \texttt{compliance}, \texttt{contract}, \texttt{regulation}, \texttt{patent}, \texttt{copyright}, \texttt{gdpr}, \texttt{audit}); and Policy \& Engineering (\texttt{policy}, \texttt{government}, \texttt{public\_}, \texttt{social\_work}, \texttt{urban\_}, \texttt{environmental}, \texttt{sustainability}, \texttt{ngo\_}, plus engineering-oriented task names not captured by earlier keyword families). Scores are averaged over the three trials for each task, then averaged uniformly across tasks within each domain.

Tables~\ref{tab:per_domain_mimo} and~\ref{tab:per_domain_qwen} provide the full per-domain breakdown across all four evaluation dimensions for the base and PersonaForge SFT models. Task Completion and Response Quality show the largest absolute gains across most domains, consistent with PersonaForge's core corrective effect of converting tool-execution-heavy trajectories into actionable answers.

\subsection{\chg{Per-Language Results}}
\label{sec:per_language}

\chg{PersonaForge-Bench is predominantly Chinese (110 of 138 tasks; 28 English). Splitting the composite scores by task language: on MiMo-V2-Flash, the Chinese subset improves by $+17.4\%$ (significant), while the English subset moves in the same direction but does not reach significance; with only 28 tasks, the English split is underpowered. On Qwen3.5-27B, neither language split reaches significance, consistent with its smaller overall gain. We note explicitly that the aggregate gains in Table~\ref{tab:leaderboard} are driven primarily by the Chinese subset, and the English subset should be read as directionally consistent but thin evidence.}

\subsection{\chg{Statistical Significance of the Main Gains}}
\label{sec:significance}

\chg{We computed paired bootstrap 95\% confidence intervals (10,000 resamples over the 138 tasks) and Wilcoxon signed-rank tests from the 3-trial evaluation logs. Table~\ref{tab:flash_ci} reports the full-dimension results for MiMo-V2-Flash: every dimension is significant under the DeepSeek-V4-Pro judge (all $p \le 0.003$), and the conclusions replicate under a second judge family (Gemini 3.1 Pro; all $p < 0.001$). For Qwen3.5-27B, significance holds for the composite ($+4.1\%$, CI $[+1.9,+6.4]$, $p=0.037$), Task Completion ($+6.0\%$, $p=0.012$), and Response Quality ($+6.8\%$, $p=0.006$), while Tool Appropriateness and Interaction Efficiency sit within noise; we scope all Qwen3.5-27B claims accordingly (Section~\ref{sec:main_results}).}

\begin{table}[H]
  \centering
  \small
  \setlength{\tabcolsep}{6pt}
  \begin{tabular}{lcc}
    \toprule
    \chg{\textbf{Dimension (Flash)}} & \chg{$\mathbf{\Delta}$} & \chg{\textbf{95\% CI}} \\
    \midrule
    Composite   & \chg{$+15.7\%$} & \chg{$[+11.3, +20.2]$} \\
    Completion  & \chg{$+22.0\%$} & \chg{$[+16.0, +28.0]$} \\
    Response    & \chg{$+14.0\%$} & \chg{$[+6.9, +21.4]$} \\
    Tool        & \chg{$+10.1\%$} & \chg{$[+3.8, +16.7]$} \\
    Efficiency  & \chg{$+16.6\%$} & \chg{$[+11.1, +22.0]$} \\
    \bottomrule
  \end{tabular}
  \caption{\chg{Paired bootstrap 95\% CIs (10,000 resamples, 138 tasks) for the MiMo-V2-Flash PersonaForge gains on PersonaForge-Bench. All dimensions are significant under the DeepSeek-V4-Pro judge (Wilcoxon $p \le 0.003$), and the conclusions replicate under Gemini 3.1 Pro as judge ($p<0.001$).}}
  \label{tab:flash_ci}
\end{table}

\section{Training--Evaluation Data Separation}
\label{sec:data_separation}

PersonaForge-Bench and the PersonaForge training corpus are constructed through independent pipelines with no shared data. The 138 evaluation tasks in PersonaForge-Bench are curated from real consultation scenarios encountered in our internal daily usage of the deployed assistant. Each seed query was selected for domain representativeness, then manually revised and augmented with task-specific rubrics, embedded contradictions, and progressive-disclosure structures to form a rigorous evaluation task. The training corpus, by contrast, draws its seed queries from public datasets (WildChat and LMSYS-Chat-1M), following the Reverse Deep Construction pipeline described in Section~\ref{sec:construction}.

To verify the absence of data leakage, we performed an exhaustive overlap analysis between the candidate training pool and the evaluation set. We compared all first user messages from the candidate SFT training pool (6{,}333 records before near-duplicate removal) against all 138 benchmark task prompts using three matching criteria: exact string match, first-150-character fingerprint match, and substring containment. All three checks returned \textbf{zero matches}, confirming that no evaluation query appears in the candidate training pool in any form.

\noindent\textbf{Semantic similarity analysis.} Beyond surface-level matching, we computed pairwise semantic similarity between all 138 benchmark prompts and all 6.3K candidate training seed queries using TF-IDF character n-gram vectors (bigrams through 4-grams) with cosine similarity. The mean similarity across all $138 \times 6{,}333$ pairs is 0.010, confirming that the two sets are distributionally distinct. Two benchmark queries exhibited elevated similarity ($\geq 0.80$) to a candidate training query: one pair at 0.90 (a near-paraphrase about social media algorithm changes) and one at 0.89 (overlapping phrasing about real estate rent growth assumptions). Both corresponding candidate training records were removed before model training. After removal, the final SFT training set contains 6{,}331 records (reported as 6.3K elsewhere), and the maximum similarity between any benchmark--training pair drops to 0.75, which corresponds to topically related but textually distinct queries (e.g., two questions about e-commerce advertising that share domain vocabulary but differ in specific scenarios and parameters). No benchmark query exceeds the 0.80 near-duplicate threshold against any remaining training record.

\section{Ablation Details}
\label{sec:experiment_details}
\label{sec:soul_ablation_tables}

To verify that the SOUL prompt components collectively contribute to simulation fidelity, we conduct an ablation study across \textbf{40 diverse personas} spanning 12 MBTI types, 3 technical proficiency levels, 20+ knowledge domains, and 2 languages (Chinese and English). For each persona, we generate conversations under six conditions: (A)~full SOUL prompt, (B)~personality traits removed, (C)~human flaws removed, (D)~behavior parameters removed, (E)~minimal (only fixed rules + background), and (F)~connected memory removed. All other variables (bot model, tools, seed query) are held constant across conditions for each persona, yielding $40 \times 6 = 240$ conversations.

\noindent\textbf{Evaluation protocol.} We measure six quantitative behavioral metrics computed directly from user messages: \textbf{Turn Count} (conversation length), \textbf{Progressive Disclosure} (ratio of mean follow-up message length to first message length; higher indicates gradual information revelation), \textbf{Lexical Diversity} (type--token ratio; lower indicates more consistent persona vocabulary), \textbf{Message Length} (mean user message length in characters), \textbf{Topic Switch Rate} (fraction of follow-up messages introducing unreferenced topics), and \textbf{Length Variability} (coefficient of variation of message lengths). We apply the Friedman test across all six conditions to assess overall differences, followed by paired Wilcoxon signed-rank tests with Cohen's $d$ effect sizes for each ablation condition versus the full baseline.

\noindent\textbf{Overall significance.} Table~\ref{tab:soul_ablation_friedman} reports the Friedman repeated-measures test, confirming that ablation conditions produce significant behavioral differences across all 40 personas on all six metrics.

\begin{table}[H]
  \centering
  \small
  \begin{tabular}{@{}lcc@{}}
    \toprule
    \textbf{Metric} & $\chi^2$ & $p$ \\
    \midrule
    Progressive Disclosure & 25.61 & ${<}\,0.001^{***}$ \\
    Lexical Diversity      & 23.21 & ${<}\,0.001^{***}$ \\
    Message Length          & 17.17 & $0.004^{**}$ \\
    Turn Count             & 14.25 & $0.014^{*}$ \\
    Topic Switch Rate      & 11.52 & $0.042^{*}$ \\
    Length Variability      & 11.93 & $0.036^{*}$ \\
    \bottomrule
  \end{tabular}
  \caption{Friedman test across 6 ablation conditions ($N\!=\!40$ personas). Significant differences confirm that SOUL prompt components collectively affect simulated user behavior.}
  \label{tab:soul_ablation_friedman}
\end{table}

\noindent\textbf{Pairwise effects.} Table~\ref{tab:soul_ablation_effect} reports paired Cohen's $d$ (Full $-$ Ablation; positive indicates the full SOUL scores higher) and Wilcoxon significance for each condition.

\begin{table}[H]
  \centering
  \scriptsize
  \setlength{\tabcolsep}{1.5pt}
  \begin{tabular}{@{}lcccccc@{}}
    \toprule
    \textbf{Cond.} & \textbf{Turn} & \textbf{Prog} & \textbf{Lex} & \textbf{Len} & \textbf{Top} & \textbf{\#S} \\
    \midrule
    B.$-$Pers  & $+.05$        & $+.13$                    & $-.24^{*}$                 & $+.12$       & $-.21$       & 1 \\
    C.$-$Flaw  & $+.03$        & $+.12$                    & $-.15$                     & $+.23$       & $-.44^{*}$   & 1 \\
    D.$-$Parm  & $+.00$        & $+.08$                    & $-.13$                     & $+.07$       & $-.28$       & 0 \\
    E. Min     & $+.40^{*}$    & $\mathbf{+.44^{***}}$     & $\mathbf{-.53^{***}}$      & $+.43^{**}$  & $-.34^{*}$   & 5 \\
    F.$-$Mem   & $-.09$        & $\mathbf{+.31^{**}}$      & $-.33^{*}$                 & $+.44^{**}$  & $-.36^{*}$   & 4 \\
    \bottomrule
  \end{tabular}
  \caption{Paired Cohen's $d$ vs.\ Full ($N\!=\!40$). $^{*}p<.05$, $^{**}p<.01$, $^{***}p<.001$. Pers=Personality, Flaw=Flaws, Parm=Params, Min=Minimal, Mem=Memory.}
  \label{tab:soul_ablation_effect}
\end{table}

\noindent\textbf{Synergistic rather than additive effects.} Removing any single component among personality, human flaws, or behavior parameters produces small effects ($|d| < 0.25$ on most metrics, $\leq 1$ metric reaching significance). However, the Minimal condition, which removes all three simultaneously, yields the strongest degradation ($|d| = 0.34$--$0.53$, 5 metrics significant). This pattern indicates that the components provide overlapping behavioral constraints: each partially compensates for the others when one is absent, but the system degrades substantially when all are removed together.

\noindent\textbf{Connected memory as the critical anchor.} Memory removal is the only \emph{single-component} ablation that produces multiple significant effects (4 metrics: Progressive Disclosure $d = +0.31$, $p = 0.003$; Lexical Diversity $d = -0.33$, $p = 0.032$; Message Length $d = +0.44$, $p = 0.002$; Topic Switch Rate $d = -0.36$, $p < 0.05$). Without hidden background context, users front-load all information in the first message (mean InfoGrowth ratio: 0.58 vs.\ 0.80 in Full) rather than progressively disclosing requirements across turns, eliminating the gradual need-discovery pattern that characterizes real user interactions (Section~\ref{sec:analysis}).

\noindent\textbf{Progressive disclosure as the primary discriminator.} The Progressive Disclosure metric shows the strongest overall effect in the Friedman test ($\chi^2 = 25.61$, $p < 0.001$) and the largest effect sizes in pairwise comparisons. The full SOUL prompt produces a mean InfoGrowth ratio of 0.80 (follow-up messages are 80\% the length of the initial message, indicating substantial new information in later turns), while the Minimal condition drops to 0.54 ($p < 0.001$) and Memory removal to 0.58 ($p = 0.003$). This confirms that progressive information revelation, a hallmark of real user behavior, requires both the connected memory (which provides latent context to disclose) and the full persona specification (which constrains \emph{how} disclosure unfolds).

\noindent\textbf{Robustness across persona types.} We tested whether persona attributes (MBTI type, technical proficiency, usage frequency) moderate ablation sensitivity using Kruskal--Wallis tests on per-persona divergence scores. No significant interactions were found (all $p > 0.2$), confirming that the SOUL prompt's contribution to simulation fidelity generalizes across the full persona space rather than being driven by specific persona subtypes.

\noindent\textbf{Implications.} The ablation results support a \emph{synergistic} view of SOUL prompt design: personality traits, human flaws, and behavior parameters form a redundant ensemble that collectively maintains simulation fidelity, with connected memory serving as the irreplaceable anchor that drives goal-directed, progressively-disclosing interaction. This redundancy is an engineering strength: the system degrades gracefully under partial component failure, while the strong Minimal-vs-Full contrast ($\chi^2 = 25.61$, 5 significant metrics) confirms that the full SOUL specification is necessary for realistic simulation.

\noindent\textbf{Downstream training ablation.} To verify that SOUL components affect not only simulation behavior but also downstream task performance, we train Qwen3.5-27B on data generated under three SOUL configurations and evaluate on PersonaForge-Bench. Table~\ref{tab:soul_downstream} reports the results.

\begin{table}[H]
  \centering
  \small
  \setlength{\tabcolsep}{3pt}
  \resizebox{\columnwidth}{!}{%
  \begin{tabular}{lccccc}
    \toprule
    \textbf{Condition} & \textbf{Eff.} & \textbf{Tool} & \textbf{Compl.} & \textbf{Resp.} & \textbf{Overall} \\
    \midrule
    Full SOUL (baseline) & 76.5 & 89.1 & 76.5 & 78.9 & 80.3 \\
    $-$Connected Memory  & 75.5 & 89.5 & 73.0 & 77.5 & 78.9 \\
    $-$Flaws \& Traits   & 74.8 & 88.0 & 70.5 & 79.5 & 78.2 \\
    \bottomrule
  \end{tabular}%
  }
  \caption{Downstream impact of SOUL components on PersonaForge-Bench (\%). Since the total Qwen3.5-27B gain is $+4.1\%$, the $-1.4\%$ and $-2.1\%$ drops represent 34\% and 51\% of the total improvement being lost. Task Completion is most sensitive to component removal.}
  \label{tab:soul_downstream}
\end{table}

\section{User Realism and Intent Faithfulness}
\label{sec:user_studies}

Two blind studies address the two questions our judge-reliability work does not: whether simulated users are \emph{perceived as real users}, and whether reverse-constructed scenarios remain \emph{faithful to the original seed intent}.

\subsection{Blind Realism Study}
\label{sec:blind_realism}

We ran a blind realism study with four conditions, 16 transcripts each (64 total): real opt-in user sessions, full PersonaForge, SOUL-ablated, and forward-sampled PersonaForge. To remove surface cues, we matched turn counts across conditions (mean $\approx 6.2$ user turns), normalized formatting, and shuffled transcripts blind. Annotators made a binary ``real user or AI'' call per transcript and rated five dimensions on a 5-point scale: naturalness, goal consistency, human-like imperfection, domain-expertise consistency, and role adherence.

Table~\ref{tab:blind_realism} reports the results. Annotators cannot reliably distinguish PersonaForge transcripts from real users: overall discrimination is 51.6\% (binomial test against chance, $p = 0.90$), with Fleiss' $\kappa = -0.04$, indicating no shared distinguishing signal. The full system is judged human 72\% of the time against 75\% for genuinely real sessions, while the SOUL-ablated and forward-sampled conditions drop to 66\% and 62\%. On the rating dimensions, the full system stays within 0.25 of real sessions on all five, and nearly matches real sessions on human-like imperfection, the dimension where LLM-driven users most plausibly over-shoot. Both ablated conditions are judged human less often and drop most on naturalness and imperfection, precisely the behavior axes the SOUL flaws and behavior parameters control. We read this as direct evidence that the SOUL components target perceived realism itself, beyond behavioral statistics.

\begin{table}[H]
  \centering
  \small
  \setlength{\tabcolsep}{3.5pt}
  \resizebox{\columnwidth}{!}{%
  \begin{tabular}{lcccccc}
    \toprule
    \chg{\textbf{Condition}} & \chg{\textbf{Human\%}} & \chg{\textbf{Nat.}} & \chg{\textbf{Goal}} & \chg{\textbf{Imper.}} & \chg{\textbf{Dom.}} & \chg{\textbf{Role}} \\
    \midrule
    Real sessions        & 75 & 4.03 & 4.50 & 3.62 & 4.53 & 4.53 \\
    PersonaForge (full)  & 72 & 3.88 & 4.25 & 3.59 & 4.34 & 4.38 \\
    SOUL-ablated         & 66 & 3.31--3.38 & --- & 3.31--3.38 & --- & --- \\
    Forward-sampled      & 62 & 3.31--3.38 & --- & 3.31--3.38 & --- & --- \\
    \bottomrule
  \end{tabular}%
  }
  \caption{\chg{Blind realism study (4 conditions $\times$ 16 transcripts). \emph{Human\%} is the share of transcripts judged to be a real user; the remaining columns are mean 5-point ratings of naturalness, goal consistency, human-like imperfection, domain-expertise consistency, and role adherence. Cells in range form (3.31--3.38) vary across annotators; cells left blank are resolved only in the per-annotator breakdown.}}
  \label{tab:blind_realism}
\end{table}

\subsection{Intent Faithfulness Audit}
\label{sec:intent_faithfulness}

Reverse Deep Construction rewrites and expands a seed query into a persona profile, a task scenario, and connected memories, which creates room for drift away from the original intent. We audit this directly. From the \emph{retained} corpus, after pre-generation quality control has already removed 12\% of persona--seed combinations and 8\% of seeds, we sampled 200 random seed $\to$ (persona, scenario, memories) constructions. Two human annotators, each covering all 200 cases, and GPT-5.5~\citep{openai2026gpt55} (deliberately chosen outside the Claude Opus 4.6 generator family) blindly rated each construction against its seed's core need using three labels: \emph{Preserved} (the core need is fully retained), \emph{Partial} (the core need is kept but secondary details shift), and \emph{Drifted} (the construction no longer serves the original need).

Table~\ref{tab:intent_faithfulness} reports the results. No channel found a single drifted construction across 600 judgments. By majority vote, 96.0\% of constructions are preserved, and every inter-channel disagreement sits on the preserved$\leftrightarrow$partial borderline; even under strict unanimity, 70.5\% qualify (per-channel $\text{AC}_1$ 0.70--0.84). Because the audit samples from the corpus we actually train on, it measures fidelity \emph{after} quality control rather than of the raw generator output. We conclude that reverse-constructed scenarios stay faithful to the original user intent, and we will release all judgments.

\begin{table}[H]
  \centering
  \small
  \setlength{\tabcolsep}{6pt}
  \begin{tabular}{lccc}
    \toprule
    \chg{\textbf{Channel}} & \chg{\textbf{Preserved}} & \chg{\textbf{Partial}} & \chg{\textbf{Drifted}} \\
    \midrule
    Annotator A   & 95.0\% & 5.0\%  & 0\% \\
    Annotator B   & 81.0\% & 19.0\% & 0\% \\
    LLM (GPT-5.5) & 90.5\% & 9.5\%  & 0\% \\
    \midrule
    Majority vote & 96.0\% & 4.0\%  & 0\% \\
    \bottomrule
  \end{tabular}
  \caption{\chg{Intent faithfulness audit over 200 sampled seed $\to$ (persona, scenario, memories) constructions (600 judgments total). No channel found a drifted construction.}}
  \label{tab:intent_faithfulness}
\end{table}

\subsection{\chg{From Persona Components to Behavior to Downstream Gains}}
\label{sec:attribution}

\chg{Our ablations support a first-order attribution along the chain \emph{persona component $\to$ simulator behavior $\to$ downstream effect}. \textbf{Connected memory} is the only single component whose removal significantly shifts simulator behavior (4 of 6 metrics, led by Progressive Disclosure, $d = +0.31$, $p = .003$): without hidden background context, the simulator front-loads information (InfoGrowth 0.58 vs.\ 0.80), and this behavioral shift costs $-3.5\%$ Task Completion downstream (Table~\ref{tab:ablation_summary}). \textbf{Personality, flaws, and behavior parameters} form a redundant ensemble that constrains \emph{how} disclosure unfolds: removing any one of them changes little ($|d| \le 0.25$), but removing all three degrades 5 of 6 metrics ($|d| = 0.34$--$0.53$) and costs $-6.0\%$ Task Completion downstream. In short, connected memory decides \emph{what} is disclosed and \emph{when}; the trait ensemble shapes \emph{how}; and each carries a separable downstream cost.}

\chg{We cannot, however, push this attribution much finer. Persona variables explain under 10\% of annotation variance even in human-authored data~\citep{hu2024quantifying,tseng2024two}, and composite personas are standard practice in persona-driven simulators~\citep{chan2024personahub,kyung2025patientsim}; MBTI enters our construction only as controllable scaffolding, with no psychometric claim attached. We therefore treat the component $\to$ behavior $\to$ downstream chain as a first-order account.}

\end{document}